\documentclass[lettersize,journal,compsoc]{IEEEtran}
\usepackage{amsmath,amsfonts}
\usepackage{amssymb}
\usepackage{bm}
\usepackage{algorithmic}
\usepackage{algorithm}
\usepackage{array}
\usepackage[font=small]{caption}
\usepackage{textcomp}
\usepackage{stfloats}
\usepackage{url}
\usepackage{verbatim}
\usepackage{enumitem}
\usepackage{dsfont}
\ifCLASSINFOpdf
  \usepackage[pdftex]{graphicx}
\else
  \usepackage[dvips]{graphicx}
\fi
\usepackage{subcaption}
\usepackage{cite}
\usepackage{xcolor,colortbl}
\definecolor{refcolor}{rgb}{0.60,0.21,0.21}
\usepackage[colorlinks,allcolors=refcolor]{hyperref}
\newcommand\sotaa{\textbf}
\newcommand\sotab{\underline}

\definecolor{Gray}{rgb}{0.95, 0.95, 0.95}
\usepackage{multirow}
\usepackage{booktabs}
\begin{document}

\title{ATD: Improved Transformer with \\Adaptive Token Dictionary for Image Restoration}

\author{Leheng Zhang, Wei Long, Yawei Li, Xingyu Zhou, Xiaorui Zhao, Shuhang Gu
% ~\IEEEmembership{Staff,~IEEE,}
        % <-this % stops a space
\thanks{\hspace*{0.6em} This work was supported by National Natural Science Foundation of China (No. 62476051) and Sichuan Natural Science Foundation (No. 2024NSFTD0041).}
\thanks{\hspace*{0.6em} Leheng Zhang, Wei Long, Xingyu Zhou, Xiaorui Zhao, and Shuhang Gu are with School of Computer Science and Engineering (School of Cyber Security), University of Electronic Science and Technology of China (UESTC), China (Email: \{lehengzhang12, weilong003, xy.chous526, zzzhaoxiaorui, shuhanggu\}@gmail.com).}
\thanks{\hspace*{0.6em} Yawei Li is with the Computer Vision Lab, ETH Z\"urich, Z\"urich, Switzerland and with the Integrated Systems Laboratory, ETH Z\"urich, Z\"urich, Switzerland (Email: li.yawei.ai@gmail.com).}% <-this % stops a space
\thanks{\hspace*{0.6em} Shuhang Gu is the corresponding author.}
}

\markboth{Journal of \LaTeX\ Class Files,~Vol.~14, No.~8, August~2021}%
{Shell \MakeLowercase{\textit{et al.}}: A Sample Article Using IEEEtran.cls for IEEE Journals}


\IEEEtitleabstractindextext{
\begin{abstract}
Recently, Transformers have gained significant popularity in image restoration tasks such as image super-resolution and denoising, owing to their superior performance.
However, balancing performance and computational burden remains a long-standing problem for transformer-based architectures.
Due to the quadratic complexity of self-attention, existing methods often restrict attention to local windows, resulting in limited receptive field and suboptimal performance.
To address this issue, we propose Adaptive Token Dictionary (ATD), a novel transformer-based architecture for image restoration that enables global dependency modeling with linear complexity relative to image size.
The ATD model incorporates a learnable token dictionary, which summarizes external image priors (i.e., typical image structures) during the training process.
To utilize this information, we introduce a token dictionary cross-attention (TDCA) mechanism that enhances the input features via interaction with the learned dictionary.
Furthermore, we exploit the category information embedded in the TDCA attention maps to group input features into multiple categories, each representing a cluster of similar features across the image and serving as an attention group.
We also integrate the learned category information into the feed-forward network to further improve feature fusion.
ATD and its lightweight version ATD-light, achieve state-of-the-art performance on multiple image super-resolution benchmarks.
Moreover, we develop ATD-U, a multi-scale variant of ATD, to address other image restoration tasks, including image denoising and JPEG compression artifacts removal.
Extensive experiments demonstrate the superiority of out proposed models, both quantitatively and qualitatively. 
% Codes and models are publicly available at \url{http://github.com/LabShuHangGU/Adaptive-Token-Dictionary}.
\end{abstract}

}

\maketitle

% \IEEEpubidadjcol

\section{Introduction}
\IEEEPARstart{I}{mage} Restoration (IR), which includes tasks such as image super-resolution, image denoising, JPEG compression artifacts removal, is a fundamental problem in the low-level vision community.
It targets at reconstructing high-quality (HQ) images from their degraded low-quality (LQ) counterpart. 
Due to the severe information loss during the degradation process, IR tasks are inherently ill-posed and present significant challenges.
This challenge has led to two primary lines of research: one focusing on non-blind restoration using synthetic datasets with known degradations (e.g., bicubic downsampling~\cite{Dong_2015_srcnn, Kim_2016_vdsr, lim2017edsr, liang2021swinir, chen2023activating, li2023grl, zhang2024transcending}, Gaussian noise~\cite{zhang2017beyond, zhang2017learning, zhang2019rnan, zhang2021plug, Zamir2021Restormer}), and another tackling blind, real-world restoration. Approaches for the latter range from models trained to reverse complex degradations directly~\cite{michaeli2013nonparametric, zhang2018learning, li2022all, chen2022simple, potlapalli2023promptir, cui2024revitalizing, conde2024instructir, cui2024omni} to generative methods employing GANs~\cite{zhang2021designing, wang2021real, wei2021unsupervised, liang2022efficient} or diffusion models~\cite{saharia2022image, rombach2022high, yue2023resshift, wu2024one, ye2024learning, zhang2025uncertainty} to handle complex and unknown degradation processes.

Early deep learning based methods~\cite{Dong_2015_srcnn, Kim_2016_vdsr, zhang2017beyond, lim2017edsr, zhang2018rcan} utilize convolutional neural networks (CNNs)~\cite{krizhevsky2012imagenet, simonyan2014very} to extract image features and establish a complex end-to-end mapping between LQ and HQ images.
Compared to traditional dictionary learning based techniques~\cite{buades2005non, yang2010image, he2011single, zhang2011color, gu2014weighted}, CNNs demonstrate superior capability in handling complex and challenging scenarios, marking a significant advancement in image restoration.
However, CNNs primarily focus on learning image priors within local neighborhoods, lacking the ability to effectively model long-range dependencies or adapt to diverse input variations.
Recently, Transformers~\cite{vaswani2017attention} have emerged as a prevalent architecture in the low-level vision literature.
At their core lies the self-attention mechanism, which captures contextual information by exploiting self-similarities throughout the input image.
By leveraging the redundancy of recurring patterns in natural images~\cite{glasner2009super, zontak2011internal}, transformers facilitate effective modeling of long-range dependencies.
In this manner, transformer-based methods support content-aware feature extraction and achieve superior performance across various image restoration tasks~\cite{liang2021swinir, Zamir2021Restormer, chen2023activating, li2023grl, li2024hierarchical, ma2024pretraining, zhang2024transcending}.

\begin{figure*}
    \centering
    \includegraphics[width=\linewidth]{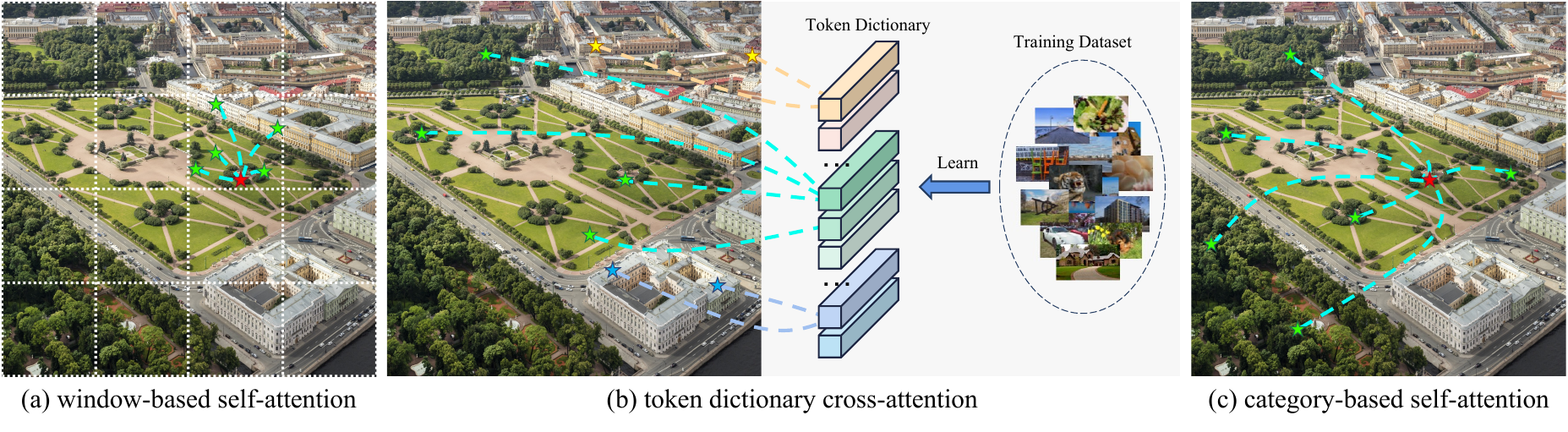}
    % \vspace{-3mm}
    \caption{Visual comparisons on different attention mechanisms. (a) Window-based self-attention constrain the attention to local windows due to the quadratic computational complexity, resulting in limited receptive field. (b) Our proposed token dictionary cross-attention leverages the typical image structures summarized in the token dictionary to incorporate external information to input features. (c) Our proposed adaptive category-based self-attention exploits relationships within categories, connecting distant but similar tokens across the image. }
    \label{fig:teaser}
    % \vspace{-3mm}
\end{figure*}

Despite recent advances through architectural refinements of transformers, several challenges remain unresolved.
One of the most critical issues is how to achieve better balance between computational cost and reconstruction performance.
Earlier approaches~\cite{zhang2019rnan, Mei2020image} introduce non-local attention modules into CNNs to improve performance, but they incur substantial computational overhead, particularly when processing high-resolution inputs.
This problem becomes even more pronounced in transformers, which rely heavily on self-attention layers.
Due to the quadratic computational complexity of self-attention, it is typically restricted to local windows~\cite{wang2022uformer, liang2021swinir}, resulting in limited receptive fields and suboptimal performance.
\cite{chen2023activating} shows that enlarging windows can improve self-attention by enabling broader self-similarity mining.
However, this improvement comes at the cost of significantly reduced efficiency and practicality.
To mitigate this issue, several studies~\cite{zhou2023srformer, li2023grl} have proposed dimensionality reduction method to support larger windows.
Nevertheless, these approaches still rely on dense interactions within local neighborhoods, which inherently presents challenges in balancing performance and efficiency.
In parallel, another line of work explores sparse self-attention mechanisms~\cite{zhang2023accurate, omni_sr} in an atrous manner.
While these approaches enable global interactions with affordable complexity, they may suffer from poor relevance preservation among the sparsified attention elements, compared to the densely connected local windows.
Consequently, it remains a fundamental challenge to facilitate meaningful global interactions without deviating from the constraint of linear computational complexity.

% \IEEEpubidadjcol

In this paper, we propose novel attention mechanisms to address the aforementioned limitations of window-based self-attention mechanism and its variants, leading to the \textbf{A}daptive \textbf{T}oken \textbf{D}ictionary (\textbf{ATD}) model for image restoration tasks.
\textbf{First}, inspired by traditional dictionary learning techniques~\cite{elad2006image, dabov2007image, yang2010image, zeyde2012single}, we establish an auxiliary token dictionary to incorporate external information.
We then introduce token dictionary cross-attention (TDCA), which operates between the input image and the token dictionary, allowing the model to exploit representative image structures learned during training.
To emulate the sparsity characteristic of traditional dictionary learning, we incorporate a logarithmic scaling strategy, which constrains the TDCA to selectively attend to the most relevant tokens while diminishing the influence of unrelated ones.
\textbf{Second}, we propose to partition the image feature based on their mutual similarity, instead of spatial coordinates (window-based partition).
The proposed partition strategy divides image feature into distinct categories, where each image token is grouped according to the index of its most relevant dictionary token.
In this way, each category aggregates structurally similar features across the image, facilitating global interactions beyond local windows while maintaining linear computational complexity.
\textbf{Lastly}, we enhance the feed-forward network (FFN) by incorporating the category information of each image token, leading to the proposed Category-aware FFN (CFFN).
The categorical embedding derived from the most relevant dictionary token is concatenated with the corresponding image token, which allows the FFN to condition its transformation on category information. 
It enables a more adaptive fusion of image features with the token dictionary and boosts restoration performance of ATD.

In summary, contributions of this work are three-fold:
\begin{itemize}
    \item We establish a novel transformer-based framework for image restoration tasks. It incorporates a learnable token dictionary which leverages the external information from training datasets. Moreover, it supports a content-aware category-based partition strategy, which enables global self-attention with linear complexity to the input size.

    \item We employ a reparameterization scheme on the scaling factor of token dictionary cross-attention to alleviate the attention weight dilution issue. Then we introduce a category-aware FFN to adaptively aggregate image features with the token dictionary. The proposed architectural modifications further improve the representation capability.

    \item We develop the Adaptive Token Dictionary (ATD) model and its multi-scale variant ATD-U, based on the proposed attention mechanisms. Extensive experiments show that ATD models achieve state-of-the-art results on both synthetic and real-world image restoration benchmarks, demonstrating the effectiveness of global dependency modeling.
\end{itemize}

A preliminary version of this work has been published in CVPR 2024~\cite{zhang2024transcending}, where the proposed approach was initially applied to image super-resolution tasks.
Building upon this foundation, the present extended study introduces several substantial improvements.
Firstly, it enhances the architecture by reparameterizing the scaling factor in the TDCA branch, mitigating the issue of attention weight dilution.
Secondly, we incorporate category information into the FFN, thereby enabling adaptive feature fusion and further improving the representational capacity of ATD.
Thirdly, we provide additional experimental results and analyses for a better understanding of the proposed framework. These include evaluations of computational complexity and convergence properties, detailed comparisons with existing sparse attention methods, and visualizations of the proposed categorization operation.
Furthermore, we develop a U-Net-based variant, referred to as ATD-U, to validate the effectiveness of the proposed attention mechanisms in multi-scale architectures.
It broadens the scope of the ATD framework to address a wider range of image restoration tasks, including image denoising and JPEG compression artifacts removal.
Extensive experiments validate the effectiveness of the proposed enhancements and the versatility of ATD framework across different restoration scenarios.

\section{Related Works}
\subsection{Example-based Image Restoration}
Before the advent of deep learning, example-based strategies are among the most widely adopted approaches for image restoration.
These methods can be broadly categorized into two main directions based on the source of exemplars.
The first group focuses on exploiting internal self-similarity within a single image~\cite{buades2005non, buades2008nonlocal, mairal2009non, glasner2009super, freedman2011image, huang2015single} to guide the reconstruction process.
For instance, \cite{buades2005non} propose the non-local means algorithm, which computes the denoised value by performing a weighted average over pixels with similar neighborhoods.
\cite{glasner2009super} utilizes multi-scale patch recurrence within the image to produce high-resolution estimates from a single input image.
\cite{freedman2011image} further constrain the matching space of example patches to highly localized regions and design taliored filter banks for improved efficiency and restoration quality.
The second group of methods derives priors from external datasets~\cite{chang2004super, elad2006image, dabov2007image, yang2010image, kim2010single, zeyde2012single, bevilacqua2012low, yang2012coupled, timofte2013anchored, gu2015convolutional, timofte2015a}. 
These studies typically involves learning paired low-quality and high-quality image patches,  using external exemplars to model the relationship between LQ and HQ images.
Representative techniques include dictionary learning~\cite{elad2006image, yang2010image}, anchored neighborhood regression~\cite{timofte2013anchored, timofte2015a}, and convolutional sparse coding~\cite{gu2015convolutional} are proposed to further enhance the mapping between LQ and HQ images and image restoration performance.

\subsection{CNN-based Image Restoration}
Following the success of convolutional neural networks in high-level vision tasks~\cite{krizhevsky2012imagenet, simonyan2014very, szegedy2015going, he2016deep}, a series of studies~\cite{Dong_2015_srcnn, dong2015compression, zhang2017beyond} demonstrate the superiority of CNN architectures over carefully designed traditional methods in image restoration tasks.
In these approaches, external prior information is implicitly encoded in the learned network weights, facilitating effective image reconstruction.
Subsequent research has explored a broad spectrum of architectural innovations to further enhance performance.
Residual learning are adopted in \cite{Kim_2016_vdsr, zhang2017learning}, while deeper architectures based on residual-in-residual and residual dense blocks are designed in \cite{lim2017edsr, Zhang_2018_rdn} to further stabilize the training process.
The effectiveness of recursive architectures~\cite{kim2016deeply, liu2018non} and graph neural networks~\cite{zhou2020cross} has also been investigated.
Meanwhile, several methods apply cascading mechanism~\cite{Ahn_2018_carn}, information multi-distillation~\cite{Hui_2019_imdn}, and lattice block~\cite{Luo_2020_latticenet} to construct lightweight framework.
\cite{zhang2022heat, zhang2022fluid} leverage the relevance between the SR process and specific physical priors (heat transfer and fluid dynamics), devising improved residual structures based on high-order finite difference equation.
In addition to the improvements in convolutional architectures, several studies have introduce attention modules to augment representation ability. 
Channel attention modules~\cite{zhang2018rcan, Dai_2020_san, Niu_2020_han, chen2022simple} exploit first- and second-order inter-channel dependencies. 
Non-local attention~\cite{zhang2019rnan} enables modeling long-range spatial correlations, while \cite{Mei2020image} further extends it with cross-scale feature interaction.
Despite their effectiveness, these attention-based modules generally exhibit quadratic computational complexity relative to the input size, making them impractical for high-resolution inputs.
To alleviate this issue, \cite{Mei_2021_nlsa} introduces a hashing strategy to gather related features in each attention bucket, therefore sparsifying the non-local attention operation and significantly improving efficiency.
These studies preliminarily validate the efficacy of attention in CNNs, leading to the development of Transformer-based methods.

\subsection{Transformer- and Mamba-based Image Restoration}
\label{sec:transformer}
In the field of natural language processing, the vanilla transformer~\cite{vaswani2017attention} leverages the self-attention mechanism to model relationships among words and capture contextual dependencies.
Inspired by this success, ViT~\cite{Dosovitskiy_2020_vit} extends the Transformer architecture to vision tasks by treating each image patch as a token, effectively modeling spatial relations between patches.
Subsequently, exploring the design space of vision transformers~\cite{touvron2021training, wang2021pyramid, wang2021not, heo2021rethinking, chu2021twins, chen2021crossvit, yang2022lite}, particularly through modifications of the self-attention mechanism, has become a major trend in the computer vision community.
However, in image restoration tasks, patch-level representations are suboptimal for reconstructing fine-grained details.
To this end, Uformer~\cite{wang2022uformer} integrates U-Net with pixel-level self-attention computed within $8 \times 8$ local windows, enhancing capability for local context modeling and detail recovery.
Restormer~\cite{Zamir2021Restormer} incorporates transposed attention mechanism and gated FFN to improve efficiency and performance for high-resolution image restoration.
SwinIR~\cite{liang2021swinir} develops a window-shifting strategy to facilitate cross-window interactions, while SCUNet~\cite{zhang2023practical} further combines it with convolutional layer to enhance local modeling ability and designs a novel data synthesis
pipeline to address practical blind image denoising.
ART~\cite{zhang2023accurate} and OmniSR~\cite{omni_sr} investigate dilated partition strategy to achieve sparse attention, effectively modeling sparse global interaction.
Rkformer~\cite{zhang2022rkformer} utilizes a reservoir structure that sparsely connects with input neurons via random connection to improve computational efficiency. ESSAformer~\cite{zhang2023essaformer} applies spectral correlation coefficient of the spectrum to inject inductive bias in self-attention for hyperspectral images.
Xformer~\cite{zhang2023xformer} construct a dual-branch U-Net to jointly utilize spatial-wise and channel-wise feature relationships.
HAT~\cite{chen2023activating} expands attention windows to $16^2$ and apply channel attention to enlarge receptive field.
In addition, methods like SRFormer~\cite{zhou2023srformer} and GRL~\cite{li2023grl} reduce the dimensionality of attention computation, enabling efficient self-attention within larger windows (up to $32^2$).
Complementarily, PFT~\cite{long2025progressive} improves efficiency by progressively concentrating attention on the most informative tokens, thereby reducing complexity without compromising performance.
Another line of work focuses on adapting the selective state space model (Mamba)~\cite{gu2024mamba} to image restoration.
MambaIR~\cite{guo2024mambair} applies a 2D selective scan module accompanied by channel attention to address challenges of local pixel forgetting and channel redundancy. 
VMambaIR~\cite{shi2025vmambair} introduces the omni selective scan mechanism to enhance the modeling of channel dimension. 
MambaIRv2~\cite{guo2024mambairv2} further proposes an attentive mechanism to replace sequential modeling with non-causal processing among semantically relevant tokens, eliminating the need for multidirectional scanning.

\section{Motivation}
Transformers building upon window-based self-attention have shown impressive performance in image restoration tasks. 
Expanding the receptive field by enlarging the attention window~\cite{chen2023activating, li2023grl, long2025progressive} prove to be effective for improving the performance of window-based self-attention.
However, such improvements often come at the cost of increased computational complexity.
To address this trade-off, we revisit traditional dictionary-learning-based approaches for image restoration tasks, highlighting the parallels between dictionary-based representations and the attention mechanism in Transformers.
Based on these insights, we propose enhanced attention mechanisms that leverage an auxiliary token dictionary to achieve improved performance with manageable computational overhead.

\subsection{Dictionary learning for image restoration} 
\label{sec:motivation}
Dictionary learning constitutes a representative class of traditional image restoration methods~\cite{elad2006image, yang2010image, yang2012coupled}.
In the training phase, a pair of over-complete dictionaries, $\bm{D}_H, \bm{D}_L\in\mathbb{R}^{d\times K}$, is learned to represent HQ and LQ image patches.
Each pair of corresponding dictionary atoms, $\bm{\delta}_H^i, \bm{\delta}_L^i \in \mathbb{R}^{d\times 1}$, captures the mapping between a specific structural pattern and its degraded counterpart.
During the reconstruction stage, a degraded input signal $\bm{y} \in \mathbb{R}^{d}$ is first decomposed as a sparse linear combination of the LQ dictionary entries:
\begin{equation}
\label{eq:dictlq}
    \bm{y} = \bm{D}_{L}\bm{\alpha},
\end{equation}
where the coefficient vector $\bm{\alpha} \in \mathbb{R}^{K\times 1}$ is obtained by solving the following optimization problem:
\begin{equation}
    \operatorname{arg} \operatorname{min}_{\bm{\alpha}} \left \| \bm{D}_{L}\bm{\alpha} - \bm{y} \right \|_2^2 + \lambda \left \| \bm{\alpha} \right \|_1 ,
\end{equation}
with $\lambda$ being a hyper-parameter that balances sparsity and reconstruction quality.
The signal is then reconstructed by linearly combining the HQ dictionary entries according to $\bm{\alpha}$:
\begin{equation}
\label{eq:dicthq}
    \tilde{\bm{x}} = \bm{D}_{H}\bm{\alpha}.
\end{equation}

\subsection{Similarity between the computation of dictionary learning and attention mechanism}
\label{sec:similarity}
Recently, transformer-based approaches have gained widespread attention in the computer vision community.
The core component of these models, the self-attention mechanism, computes pairwise similarities between tokens and uses the resulting similarity map to aggregate information:
\begin{equation}
\label{eq:self-attention}
    \operatorname{Attention}(\bm{Q}, \bm{K}, \bm{V}) = \operatorname{SoftMax}\left( \bm{Q}\bm{K}^T/\sqrt{d} \right) \bm{V},
\end{equation}
where $\bm{Q}\in \mathbb{R}^{N\times d}$, $\bm{K}\in \mathbb{R}^{N\times d}$ and $\bm{V}\in \mathbb{R}^{N\times d}$ denote the query, key, and value matrices, obtained by applying linear transformations to the input feature $\bm{X}\in \mathbb{R}^{N\times d}$.
To some extent, self-attention shares similarities with dictionary learning, as both involve coefficient computation and linear combination.
Firstly, both methods derive weights based on the relationship between data points.
Self-attention computes pairwise dot products $\bm{A} = \operatorname{SoftMax} (\bm{Q}\bm{K}^T/\sqrt{d})$ among tokens as attention weights. 
Meanwhile, the sparse coefficients in dictionary learning are obtained by solving the optimization problem $\bm{\alpha} = \operatorname{arg}\operatorname{min}_{\bm{\alpha}} [\left \| \bm{D}_{L}\bm{\alpha} - \bm{y} \right \|_2^2 + \lambda \left \| \bm{\alpha} \right \|_1]$.
Secondly, they both utilize weighted combination for reconstruction to emphasize the most important features, i.e., $\tilde{\bm{X}} = \bm{AV}$ and $\tilde{\bm{x}} = \bm{D}_H\bm{\alpha}$.
These structural parallels allow us to draw an analogy: the query token $\bm{Q}$ and key token $\bm{K}$, and value token $\bm{V}$ correspond to the input $\bm{y}$, LQ dictionary $\bm{D}_L$, and HQ dictionary $\bm{D}_H$, respectively.
This perspective forms the basis for our proposed token dictionary attention mechanisms, which we elaborate on in the following sections.

\subsection{Advanced Cross\&Self-Attention based on Token Dictionary}
\label{sec:motivation_tdca_acmsa}
In vanilla transformers, the self-attention mechanism focuses on exploiting internal self-similarities for feature enhancement, thereby lacking the capability to explicitly model external priors.
As discussed in the previous subsection, we visit the mathematical connection between dictionary learning and attention mechanism, which offers insights for incorporating external priors into transformer architectures.
Motivated by this, we propose a learnable token dictionary to explicitly capture prior information during training.
We then introduce the token dictionary cross-attention (TDCA) mechanism, which leverages the similarity between input tokens and the learned token dictionary to effectively integrate externel priors, as detailed in Sec.~\ref{sec:TDCA}.
Building upon these external priors, we further enhance the modeling of internal similarities.
Most existing methods rely on window-based self-attention, which limits the modeling of self-similarities to local neighborhoods.
It is generally sufficient for smooth regions, where patterns tend to densely recur within their immediate vicinity.
However, such limited receptive fields fall short in capturing dependencies in more complex regions, where similar structures may be sparsely distributed across the image~\cite{zontak2011internal}.
To bridge these distant but structurally similar regions, we exploit the attention map generated by TDCA, which reveals the correlation between each query token and various structural types stored in the token dictionary.
Based on this correlation, the input can be partitioned into categories, allowing us to apply a category-based self-attention mechanism, as described in Sec.~\ref{sec:ACMSA}.
These novel strategies enhances both external and internal priors modeling, enabling global dependencies modeling with linear computational complexity and effectively overcoming the limitations of existing window-based approaches.

\begin{figure*}
    \centering
    \includegraphics[width=\linewidth]{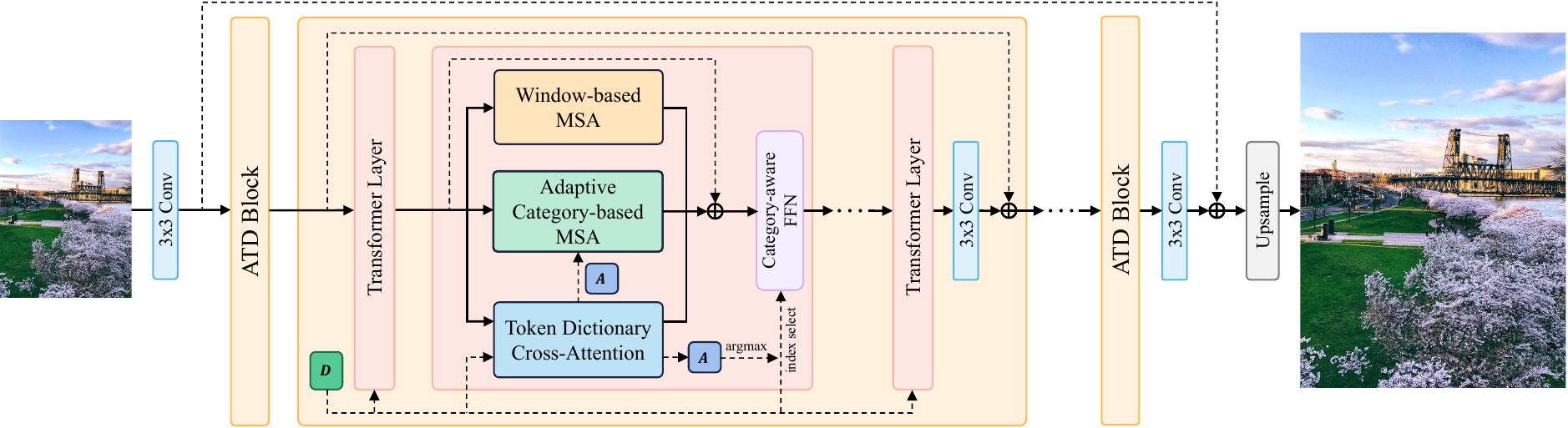}
    \vspace{1mm}
    \includegraphics[width=\linewidth]{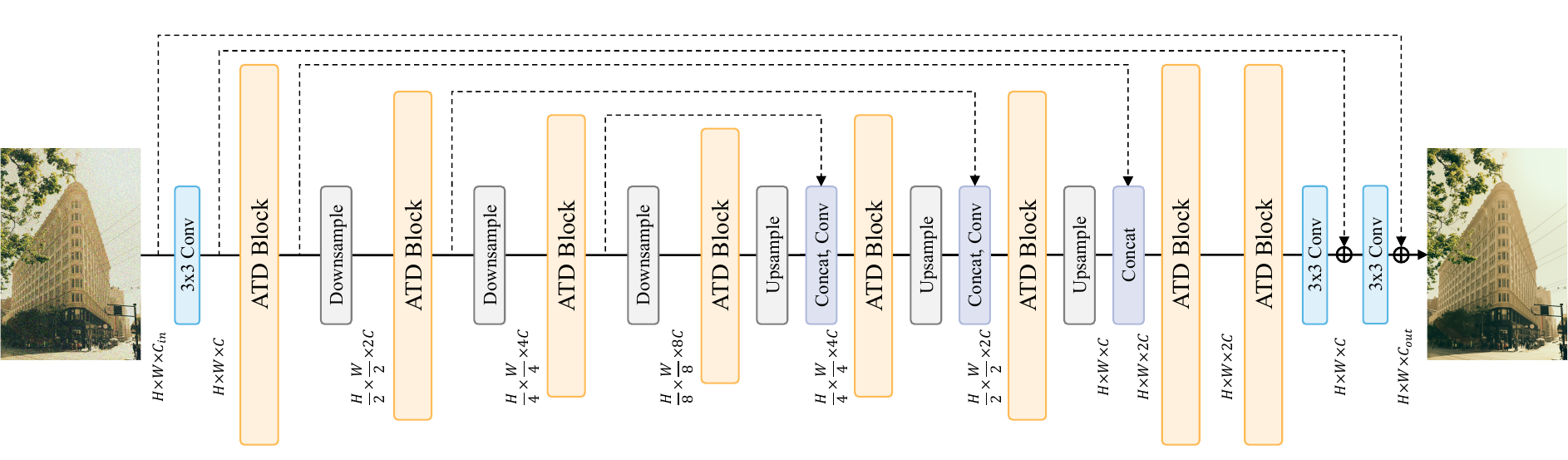}
    \vspace{-1mm}
    
    \caption{Overall architectures of the proposed ATD and ATD-U networks. Each ATD block contains several consecutive transformer layers and a learnable token dictionary. The transformer layer combines three attention branches: token dictionary cross-attention (\ref{sec:TDCA}), adaptive category-based self-attention (\ref{sec:ACMSA}), and window-based self-attention to enhance image feature. The attention map of TDCA branch is further utilized by the ACMSA branch for categorization and CFFN for dictionary entry selection, respectively. }
    \label{fig:overall_architecture}
    \vspace{-2mm}
\end{figure*}

\begin{figure*}
  \centering
  \begin{subfigure}[b]{0.40\linewidth}
    \centering
    \includegraphics[width=\linewidth]{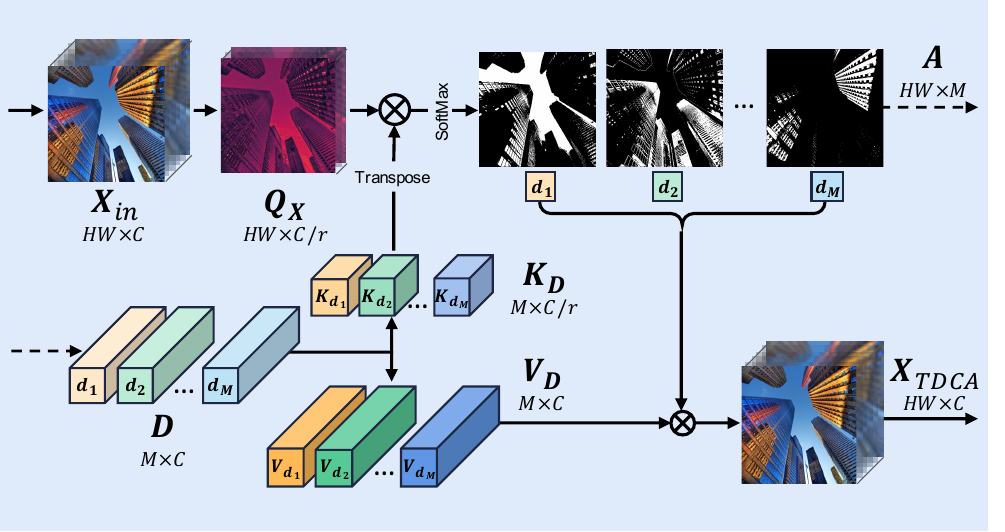}
    \caption{Token Dictionary Cross-Attention}
    \label{fig:tdca}
  \end{subfigure}%
  \hfill
  \begin{subfigure}{0.40\linewidth}
    \centering
    \includegraphics[width=\linewidth]{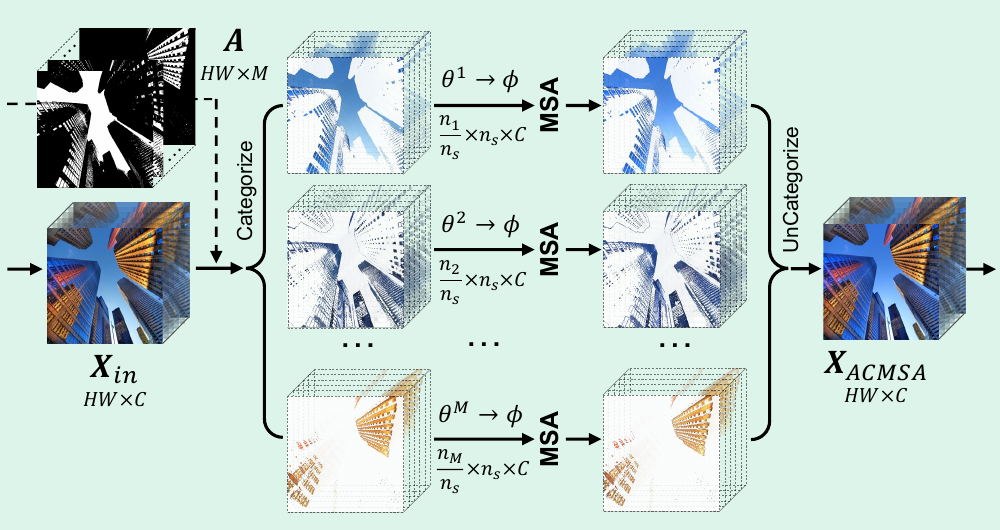}
    \caption{Adaptive Category-based MSA}
    \label{fig:acmsa}
  \end{subfigure}
  \hfill
  \begin{subfigure}{0.18\linewidth}
    \centering
    \includegraphics[width=\linewidth]{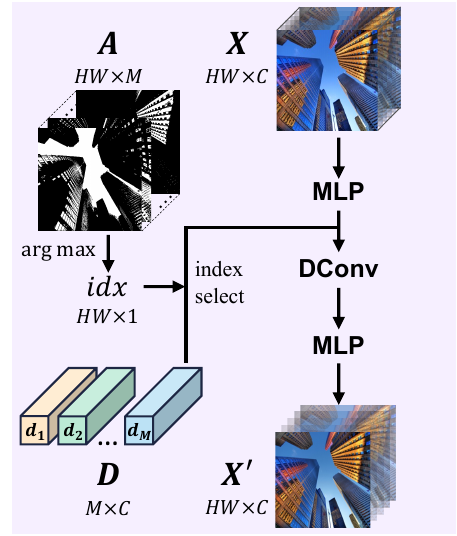}
    \caption{Category-aware FFN}
    \label{fig:cffn}
  \end{subfigure}
  % \vspace{-3mm}
  \caption{Illustrations of the proposed (a) Token Dictionary Cross-Attention (TDCA), (b) Adaptive Category-based Multi-head Self-Attention (AC-MSA), and (c) Category-aware Feed-Forward Network (CFFN). 
  Descriptions of TDCA and ACMSA are provided in Sec.~\ref{sec:TDCA} and \ref{sec:ACMSA}, respectively.
  Further details of the $\operatorname{Categorize}$ operation can be found in Eq.~\ref{eq:categorize} and Fig.~\ref{fig:categorize}.
  }
  \label{fig:illustration_tdca_acmsa}
  % \vspace{-4mm}
\end{figure*}

\section{Methodology}
Building upon the connections between attention mechanisms and traditional example-based methods, we aim to design novel cross- and self-attention mechanisms that enhance the utilization of both external and internal information.
In this section, we provide a detailed introduction to the proposed ATD model, including the overall architectures, designs of key components, and analyses of how these designs address the limitations of the existing window-based approaches.

\subsection{Network Architecture}
We establish ATD for image super-resolution and ATD-U for image denoising and JPEG artifacts removal tasks, both built upon ATD transformer layers which integrate the proposed TDCA, AC-MSA branches, and improved CFFN.
As shown in Fig.~\ref{fig:overall_architecture}, we illustrate the residual in residual architecture~\cite{liang2021swinir} of ATD, and the U-shaped architecture of ATD-U.
Both architectures comprise three stages: shallow feature extraction, deep feature extraction and image reconstruction. 
The input image $\bm{I}_L \in \mathbb{R}^{H\times W\times 3}$ is first passed through a $3\times 3$ convolutional layer $H_{SF}$ to extract shallow feature $\bm{F}_0 \in \mathbb{R}^{H\times W\times C}$ as:
\begin{equation}
    \bm{F}_0 = H_{SF}(\bm{I}_L).
\end{equation}

For image SR, the shallow feature is fed to $n_b$ ATD blocks (ATDBs) and a $3\times 3$ convolutional layer to extract deep feature:
\begin{equation}
\begin{split}
    &\bm{F}_i = H_{ATDB_i}(\bm{F}_{i-1}), i=1, 2, \cdots, n_b,\\
    &\bm{F}_{DF} = H_{DF}(\bm{F}_{n_b}),
\end{split}
\end{equation}
where each ATD block contains $n_l$ consecutive ATD transformer layers and a learnable dictionary $\bm{D}$ shared across all layers.
In each ATD transformer layer, three branches of different attention mechanism are integrated to jointly exploit external (TDCA), global (ACMSA), and local (SWMSA~\cite{liu2021swin}) information
\begin{equation}
\begin{split}
    &\bm{X}_N = \operatorname{LN}(\bm{X}_{in}),\\
    &\bm{X}_{out} = \bm{X} + \bm{X}_{TDCA} + \bm{X}_{ACMSA} + \bm{X}_{SWMSA}.
\end{split}
\end{equation}
The summation of the output of three attention modules and the shortcut of input is then entered into CFFN as described in Eq.~\ref{eq:cffn} for fusion of local feature and category information.
Additionally, we apply the commonly used LayerNorm (LN)~\cite{ba2016layer} in each attention module and FFN for stablization.
Finally, after the ATD blocks, we use a simple reconstruction block $H_{Rec}$ to transform the features into output HQ estimation $\bm{I}_H \in \mathbb{R}^{rH\times rW\times 3}$:
\begin{equation}
    \bm{I}_H = H_{Rec}(\bm{F}_0 + \bm{F}_{DF}),
\end{equation}
where $H_{Rec}$ includes pixel-shuffle operation~\cite{shi2016real} and two convolutional layers to upsample the feature.

For ATD-U, a 4-level symmetric encoder-decoder architecture is applied for deep feature extraction, with each encoder or decoder consisting of cascaded ATD blocks without residual connection and convolutional layer.
The shallow feature $\bm{F}_0$ is entered into the encoders, with spatial resolution halved and channel dimensions doubled as stage grows to obtain the low-resolution latent features.
The decoder takes the latent feature and skip connections from encoder as input to progressively recover the high-resolution deep features $\bm{F}_{DF}$.
The deep feature is further enhanced through a refinement block $H_{ref}$.
Finally, two convolutional layers $H_{skip}$ and $H_{rec}$ are applied to generate the residuals, which are then added to the shallow feature and the input image to obtain the restored image $\bm{I}_H$:
\begin{equation}
    \bm{I}_H = \bm{I}_L + H_{rec}(H_{skip}(\bm{F}_{SF}) + H_{ref}(\bm{F}_{DF})).
\end{equation}

\subsection{Token Dictionary Cross-Attention}
\label{sec:TDCA}
In this subsection, we introduce the details of the proposed token dictionary and the corresponding token dictionary cross-attention mechanism.

Unlike existing self-attention methods that primarily focus on modeling internal dependencies within the input image, we revisit traditional dictionary learning and establish an auxiliary token dictionary to explicitly summarize external information.
We establish the token dictionary $\bm{D} \in \mathbb{R}^{M\times d}$ as learnable network parameters, where $M$ and $d$ denote the number of dictionary entries and channel dimension, respectively.
We set $M$ to be much larger than $d$ to ensure that the dictionary stores a diverse and complete set of features.
Building upon the relevance between Eq.~\ref{eq:dictlq},~\ref{eq:dicthq}, and Eq.~\ref{eq:self-attention}, we propose the Token Dictionary Cross-Attention (TDCA) mechanism:
\begin{equation}
\label{eq:tdca}
\begin{split}
    &\bm{A}_{D}=\operatorname{SoftMax}(\operatorname{Sim_{cos}}(\bm{Q}_{X}, \bm{K}_{D})\tau),\\
    &\operatorname{TDCA}(\bm{Q}_{X}, \bm{K}_{D}, \bm{V}_{D}) = \bm{A}_{D} \bm{V}_{D},
\end{split}
\end{equation}
where the query token $\bm{Q}_{X} \in \mathbb{R}^{N\times d_r}$ is linearly projected from input feature $\bm{X}$, while the key and value tokens $\bm{K}_{D}\in \mathbb{R}^{M\times d_r}, \bm{V}_{D}\in \mathbb{R}^{M\times d}$ are transformed from the token dictionary $\bm{D}$.
We compute the attention map $\bm{A}_D$ using cosine similarity $\operatorname{Sim_{cos}}(\cdot)$, scaled by a learnable temperature factor $\tau$.
In light of the dictionary learning paradigm, we discard the multi-head attention design and reduce the channel dimensionality for $\bm{Q}_X$ and $\bm{K}_D$ to $d_r$ to represent LQ input and dictionary.
This dimensionality reduction also serves to alleviate computational redundancy and overhead.
The token dictionary $\bm{D}$ and its transform matrices $\bm{W}_K, \bm{W}_V$ are randomly initialized as learnable parameters to ensure the orthogonality of dictionary atoms.
During the training phase, they are updated jointly with the rest of the network via gradient descent, allowing the model to adaptively capture degradation-aware priors from the training data.
The proposed TDCA mechanism utilizes the cosine similarity to select relevant entries from the token dictionary to enhance each input token via cross-attention.
Higher values in the resulting attention map $\bm{A}_D$ correspond to dictionary key atoms that are more similar to the query token, thereby assigning greater weights to their associated value tokens during feature aggregation.
This mechanism is analogous to the sparse coefficients computation and weighted combination processes of the traditional dictionary learning.
In this vein, the most relevant external information embedded in the token dictionary is explicitly incorporated to improve the input representation.
To validate the semantic meaningfulness of these learned priors, we visualize representative dictionary tokens and their corresponding activated features in the appendix.

Within the framework of dictionary learning, enforcing sparsity is essential for reducing redundancy and achieving more accurate representations.
However, in our preliminary work~\cite{zhang2024transcending}, attention scaling factor $\tau$ was implemented as a simple learnable parameter.
This formulation becomes inadequate as the dictionary size $M$ increases, since the enlarged dictionary space includes more irrelevant entries.
These irrelevant elements dilute the attention distribution after softmax activation, weakening the contrast among attention scores.
Consequently, the attention map becomes less sparse, making it difficult for the model to effectively identify and utilize the most informative dictionary atoms, thereby hindering performance improvement.
To address this issue, we propose to adjust the scaling factor $\tau$ based on the dictionary size $M$:
\begin{equation}
    \tau' = 1 + \tau \ln(M).
\label{eq:reparameterize}
\end{equation}
Specifically, we amplify the scaling factor by $\operatorname{ln}(M)$.
This logarithmic scaling increases the separation between attention scores of highly and less correlated dictionary entries. 
As the dictionary size grows, the resulting larger $\ln(M)$ term further amplifies the contrast in attention weights.
In this way, the attenuation of attention weights for the most relevant dictionary tokens is mitigated. 
This strategy encourages the coefficients to be sparse, which is achieved using special penalty in traditional dictionary learning and crucial for reconstruction performance.
Ablation studies in Tab.~\ref{tab:ablation_dictsize} validate its effectiveness, particularly when using larger dictionaries. 

\begin{figure}
    \centering
    \includegraphics[width=.8\linewidth]{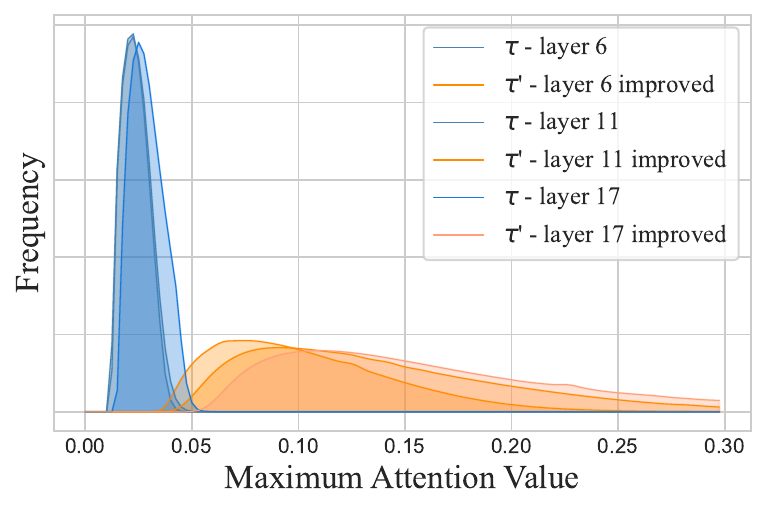}
    \caption{Distribution of the maximum attention values over image tokens with respect to the token dictionary in TDCA across different layers. Y-axis is log-scaled for better visualization.
    }
    \label{fig:vis_tdca}
\end{figure}
\noindent
\textbf{Visualization of the attention weights of TDCA.}
In Fig.~\ref{fig:vis_tdca}, we visualize the distribution of the maximum attention values between each image token and the token dictionary on the validation sets of \cite{timofte2017div2k, li2023lsdir}. 
Under the original scaling factor setting, TDCA yields less discriminative attention weights, with maximum weights often concentrated around 0.025 (close to the uniform weight $\frac{1}{M}\approx0.008$), thereby leading to suboptimal performance.
In contrast, with the proposed adjustment to the scaling factor (Eq.~\ref{eq:reparameterize}), the maximum values shift toward a broader range (0.05 to 0.30). 
This suggests that Eq.~\ref{eq:reparameterize} drives each image token to focus on the most relevant dictionary entries.
Such behavior also aligns with the idea of sparse representation, in which only a small number of dictionary tokens contribute to a given image token.

\begin{figure}
  \centering
  \includegraphics[width=1.0\linewidth]{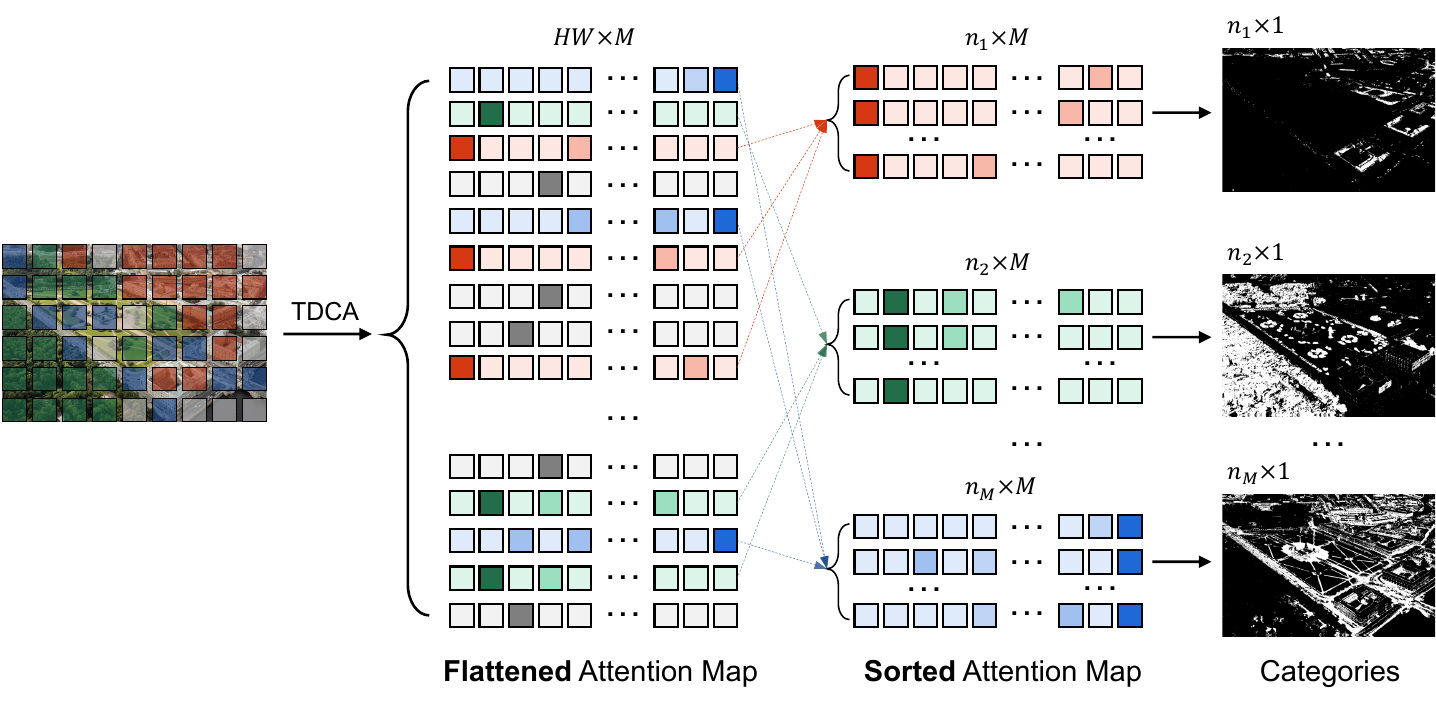}
  \caption{Illustration of the proposed $\operatorname{Categorize}$ operation.
  The attention map is flattened and sorted based on the index of the maximum value in each row as denoted in Eq.~\ref{eq:categorize} and Eq.~\ref{eq:sub-categorization}.
  This effectively clusters tokens sharing similar features for self-attention.
  Subsequently, the $\operatorname{UnCategorize}$ operation applies the inverse permutation to restore the original spatial structure.}
  \label{fig:categorize}
  % \vspace{-4mm}
\end{figure}

\subsection{Adaptive Category-based Self-Attention}
\label{sec:ACMSA}
The vanilla self-attention mechanism requires calculating pairwise similarities between all input tokens, resulting in a quadratic computational complexity with respect to input size.
This not only becomes computationally prohibitive for high-resolution inputs but also introduces redundant computations between irrelevant tokens.
A common solution is to restrict the scope of self-attention to local windows~\cite{liu2021swin, liang2021swinir}, however, this inherently limits the receptive field and constrains the modeling capacity of self-attention.
Although recent studies~\cite{chen2023activating, li2023grl, long2025progressive} demonstrate that expanding receptive field using larger windows improves performance, this strategy incurs substantial computational cost.
Moreover, window partitioning remains input-invariant, failing to eliminate redundant computations among unrelated tokens.

\begin{figure*}[h]
  \centering
  \includegraphics[width=0.9\linewidth]{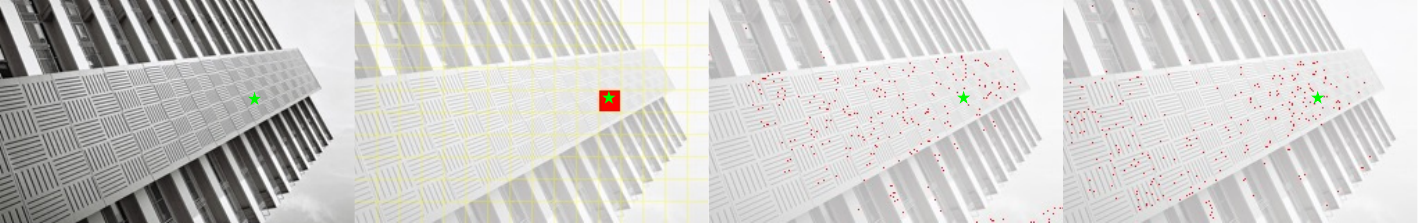}
  \includegraphics[width=0.9\linewidth]{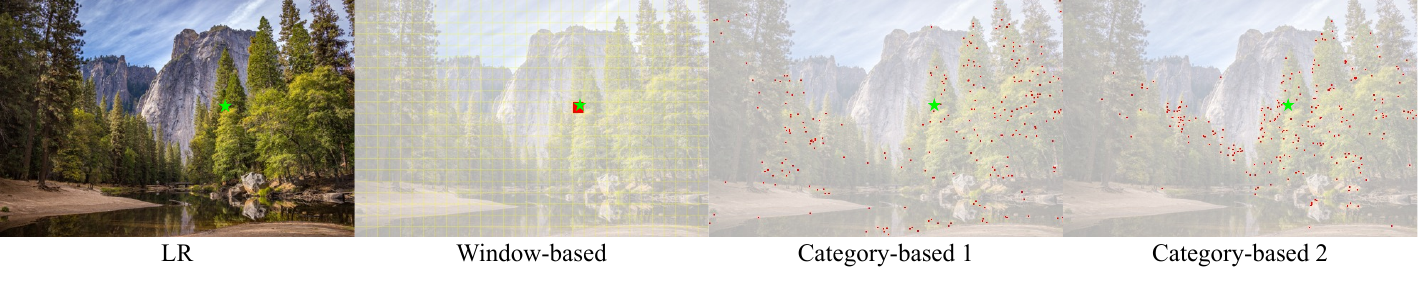}
  \caption{Visualization of the attention groups for window-based self-attention and the proposed adaptive category-based self-attention. 
  \textcolor{green}{Green} and \textcolor{red}{red} regions represent the \textcolor{green}{query tokens} and their corresponding \textcolor{red}{key tokens}, respectively. In window-based self-attention (window size $n_w=16^2$), similarity computations are confined within local windows, which limits the ability to exploit similar regions across the image.
  In contrast, the proposed adaptive category-based self-attention (category size $n_s=256$, equal to window size) could connect distant but similar tokens across the image.
  Additionally, as shown in (c) and (d), the attention categories vary across different layers, allowing the receptive field to expand progressively.
  \textit{Zoom in for a better view.}
  }
  \label{fig:visual_self_attention}
  % \vspace{-4mm}
\end{figure*}

To address these challenges, we propose a novel self-attention mechanism grounded in the token dictionary introduced in Sec.~\ref{sec:TDCA}.
We observe that the attention map produced by TDCA implicitly encodes class information for each input token.
Instead of relying on fixed 2D coordinates for window partitioning, we introduce a category-based partitioning strategy that leverages this class information:
\begin{equation}
\label{eq:categorize}
\bm{\theta}^i = \{ \bm{x}_j \mid \operatorname{arg}\operatorname{max}_k (\bm{A}_D^{jk}) = i\},
\end{equation}
where each element $\bm{A}_D^{jk}$ in the attention map $\bm{A}_D$ denotes the similarity between an input token $\bm{x}_j$ and a dictionary token $\bm{D}_k$.
If $\bm{A}_D^{ji}$ is the maximum among $\bm{A}_D^{j1}, \bm{A}_D^{j2}, \cdots, \bm{A}_D^{jM}$, we assign $\bm{x}_j$ to the $i$-th category $\bm{\theta}^i$, indicating it is most likely associated with the $i$-th dictionary token $\bm{D}_i$.
This categorization naturally enhances the internal correlation among tokens within the same group.
For any two pixels from the same group $\bm{x}_j, \bm{x}_k \in \bm{\theta}^i$, their distance can be upper-bounded by the sum of their respective distances to the dictionary token $\bm{D}_i$, according to the triangular inequality in a suitable metric space~\cite{wang2004multi, gairola2020simpropnet, li2023grl}:
\begin{equation}
\label{eq:tri}
    d(\bm{x}_j, \bm{x}_k) \le d(\bm{x}_j, \bm{D}_i) + d(\bm{D}_i, \bm{x}_k) = \bm{A}_D^{ji} + \bm{A}_D^{ki}.
\end{equation}
Although the distance metric $d(\cdot, \cdot)$ used in Eq.~\ref{eq:tdca} does not strictly satisfy this inequality, \cite{li2023grl} shows that it still supports effective similarity propagation and leads to promising results.
Therefore, under the categorization criterion defined in Eq.~\ref{eq:categorize}, the tokens $\bm{x}_j$ and $\bm{x}_k$ are expected to be similar, as their respective distances to the dictionary token, $\bm{A}_D^{ji}$ and $\bm{A}_D^{ki}$, are both minimized.

Through the above categorization strategy, we obtain $M$ categories, each comprising tokens that belong to a particular class and exhibit mutual similarity.
However, directly performing self-attention within these groups encounters challenges in parallelism and computational complexity due to the imbalance in category size.
Several categories may contain an excessive number of tokens, leading to prohibitive quadratic computational complexity.
To address this imbalance, we draw inspiration from~\cite{Mei_2021_nlsa} and further partition each category into sub-categories $\bm{\phi}$, formulated as:
\begin{equation}
\label{eq:sub-categorization}
    \begin{split}
        \bm{\phi} &= \left[ \bm{\theta}_1^1, \bm{\theta}_2^1, \cdots, \bm{\theta}_{n_1}^1, \cdots, \bm{\theta}_{n_M}^M \right], \\
        \bm{\phi}^{j} &= \left[ \bm{\phi}_{j * n_s + 1}, \bm{\phi}_{j * n_s + 2}, \cdots, \bm{\phi}_{(j+1) * n_s} \right],
    \end{split}
\end{equation}
where category $\bm{\theta}^i$ consists of $n_i$ tokens, and $n_s$ denotes the sub-category size.
We sort all tokens by their assigned category index and divide them equally into groups, forming sub-categories $\bm{\phi}^j$.
The fixed sub-category size $n_s$ facilitates parallel processing across all sub-categories.
An overview of this process is illustrated in Fig.~\ref{fig:categorize}.
We also provide visualizations of the categorization results in the appendix, which further demonstrate how the proposed method aggregates similar features and improve intra-group relevance.
Overall, the proposed Adaptive Category-based Multi-head Self-Attention (AC-MSA) operates as follows:
\begin{equation}
    \begin{split}
        \{ \bm{\phi}^{j}\} &= \operatorname{Categorize}(\bm{X}_{in}) , \\
        \hat{\bm{\phi}}^{j} &= \operatorname{MSA}(\bm{\phi}^{j} \bm{W}^{Q}, \bm{\phi}^{j} \bm{W}^{K}, \bm{\phi}^{j} \bm{W}^{V}), \\
        \bm{X}_{ACMSA} &= \operatorname{UnCategorize} (\{ \hat{\bm{\phi}}^{j} \}).
    \end{split}
\end{equation}
First, we partition input tokens through a two-stage categorization process, based on Eq.~\ref{eq:categorize} and Eq.~\ref{eq:sub-categorization}.
This operation gathers similar tokens into each sub-category, which can be viewed as forming an sparse, content-aware attention group.
Subsequently, multi-head self-attention is performed within each sub-category.
Since the categorization is based on class information rather than spatial coordinates, elements within each group are distributed across the entire image rather than confined to local neighborhoods.
Despite the constraint on sub-category size to control computational overhead, AC-MSA effectively models global dependencies, demonstrating clear advantages over window-based methods.
After the attention operation, the $\operatorname{UnCategorize}$ operation, as the inverse of $\operatorname{Categorize}$, restores each token to its original position, producing the final output feature.

\noindent
\textbf{Visualization of different self-attention mechanisms}
As illustrated in Fig.~\ref{fig:visual_self_attention}, we visualize the attention groups of different self-attention mechanisms in the context of super-resolution task.
Due to the page limit, we include visualizations on more diverse scenarios in the appendix.
For window-based self-attention (Fig.~\ref{fig:visual_self_attention} (b), with window size of $16\times 16$), the input features are partitioned into fixed square windows.
Each query token can only attend to key tokens within its local neighborhood, resulting in a limited receptive field.
Consequently, window-based self-attention struggles to leverage the information present across similar but distant regions of the image.
In contrast, the attention groups in category-based self-attention (with sub-category size $n_s = 128$) consist of tokens that are sparsely distributed throughout the image.
As shown in the first row of Fig.~\ref{fig:visual_self_attention}, structures such as building surfaces recur across multiple spatial scales, each exhibiting varying levels of degradation.
Category-based self-attention allows severely degraded regions to draw information from clearer, similar structures elsewhere in the image.
Furthermore, since each category is randomly divided into sub-categories at each layer, the attention groups vary significantly across layers.
In the second row of Fig.~\ref{fig:visual_self_attention}, where the input resolution is higher, the adavantage of our method is further pronounced.
While the receptive field of window-based methods is constrained to local neighborhoods, our category-based self-attention consistently enables global self-similarity mining without increasing computational complexity.

\subsection{Category-aware FFN}
The Feed-Forward Network (FFN) is a crucial component of transformer-based architectures, and various techniques~\cite{li2021localvit, wang2022uformer, omni_sr} have been developed to enhance FFN by better exploiting local content.
To further improve its performance, we propose a Category-aware FFN (CFFN) by incorporating the attention map from TDCA to adaptively inject category information.
Specifically, for each input token $\bm{x}_i$, we identify its most similar dictionary entry $\bm{D}_{idx_i}$ based on the TDCA attention map $\bm{A}_D$ as auxiliary guidance. 
We then concatenate the category information embedding $\bm{\delta} = \{\bm{D}_{idx_i}\}$ with the intermediate feature $\bm{X}\bm{W}_1$ before feeding it into the depth-wise convolution layer, formulated as:
\begin{equation}
\label{eq:cffn}
\begin{split}
&\ \bm{\delta}_i = \bm{D}_{idx_i}\ \ \text{where}\ \  idx_i=\operatorname{arg}\operatorname{max}_k (\bm{A}_D^{jk}),\\
&\bm{X}' = \operatorname{CFFN}(\bm{X}) = \bm{X} + \operatorname{DWConv}([\bm{X}\bm{W}_1,\bm{\delta}\bm{W}_d])\bm{W}_2.
\end{split}
\end{equation}
Through this design, category-specific information derived from TDCA is adaptively fused in local regions, thereby reinforcing the integration of external prior knowledge and enhancing the representational capability of the network.

\section{Experiments on Image Super-Resolution}
\label{sec:srexp}
\subsection{Implementation Details}
\label{sec:srdetails}
\noindent
\textbf{Network Architecture Details.}
We establish ATD and ATD-light models for classical and lightweight super-resolution tasks, respectively.
For the ATD, we set the channel dimension to $C=216$ and apply $n_b=6$ ATD blocks, each comprising $n_l=6$ ATD transformer layers.
In the attention module, the token dictionary consists of $M = 512$ entries, each initialized as a learnable tensor $\bm{d}_i\in\mathbb{R}^{1\times 216}$, forming $\bm{D}\in\mathbb{R}^{512\times 216}$.
In the TDCA branch, the reduced channel dimension is set to $d_r=20$.
In other attention branches, window size and sub-category size are set to 16 and 256 respectively, and the number of attention heads is 4.
In the lightweight version, hyperparameters are reduced to $[C, n_b, M, d_r] = [48, 4, 256, 12]$.
Additionally, the sub-category size and number of attention heads are decreased to 128 and 3.

\begin{table*}[hbtp]
\scriptsize
% \footnotesize
\setlength{\tabcolsep}{9pt}

\caption{Quantitative comparison (PSNR/SSIM) with state-of-the-art methods on \textbf{classical image SR} task. Best and second best results are marked with \sotaa{bold} and \sotab{underline}. Experimental details can be found in Sec.~\ref{sec:srdetails}. }
\vspace{-2mm}
\label{tab: results classical sr}
  % \centering
  % \begin{tabular}{l|l}
  \begin{center}
      
  \begin{tabular}{l|c|c|cc|cc|cc|cc|cc}
    \toprule
    \multirow{2}{*}{\textbf{Method}} & \multirow{2}{*}{\textbf{Scale}} & \multirow{2}{*}{\textbf{Params}} & \multicolumn{2}{c|}{\textbf{Set5}~\cite{Bevilacqua2012set5}} & \multicolumn{2}{c|}{\textbf{Set14}~\cite{zeyde2012single}} & \multicolumn{2}{c|}{\textbf{BSD100}~\cite{MartinFTM01}} & \multicolumn{2}{c|}{\textbf{Urban100}~\cite{Huang_2015_Urban100}} & \multicolumn{2}{c}{\textbf{Manga109}~\cite{Matsui_2016_Manga109}} \\

    % $\times$2 %%%%%%%%%%%%%%%%%%%%%%%%%%%%%%%%%%%%%%%%%%%%%%%%%%%
    & & & PSNR & SSIM & PSNR & SSIM & PSNR & SSIM & PSNR & SSIM & PSNR & SSIM   \\

    \midrule

    EDSR~\cite{lim2017edsr}         & $\times$2 & 42.6M & 38.11 & 0.9602 & 33.92 & 0.9195 & 32.32 & 0.9013 & 32.93 & 0.9351 & 39.10 & 0.9773 \\
    RCAN~\cite{zhang2018rcan}       & $\times$2 & 15.4M & 38.27 & 0.9614 & 34.12 & 0.9216 & 32.41 & 0.9027 & 33.34 & 0.9384 & 39.44 & 0.9786 \\
    SAN~\cite{Dai_2020_san}         & $\times$2 & 15.7M & 38.31 & 0.9620 & 34.07 & 0.9213 & 32.42 & 0.9028 & 33.10 & 0.9370 & 39.32 & 0.9792 \\
    HAN~\cite{Niu_2020_han}         & $\times$2 & 63.6M & 38.27 & 0.9614 & 34.16 & 0.9217 & 32.41 & 0.9027 & 33.35 & 0.9385 & 39.46 & 0.9785 \\
    % CSNLN~\cite{Mei2020image}       & $\times$2 & 38.28 & 0.9616 & 34.12 & 0.9223 & 32.40 & 0.9024 & 33.25 & 0.9386 & 39.37 & 0.9785 \\
    % IPT~\cite{Chen_2020_ipt}        & $\times$2 & 115M  & 38.37 & - & 34.43 & - & 32.48 & - & 33.76 & - & - & - \\
    SwinIR~\cite{liang2021swinir}   & $\times$2 & 11.8M & 38.42 & 0.9623 & 34.46 & 0.9250 & 32.53 & 0.9041 & 33.81 & 0.9433 & 39.92 & 0.9797 \\
    % EDT~\cite{li2021efficient}      & $\times$2 & 11.5M & 38.45 & 0.9624 & 34.57 & 0.9258 & 32.52 & 0.9041 & 33.80 & 0.9425 & 39.93 & 0.9800 \\
    % MambaIR~\cite{guo2024mambair}   & $\times$2 & 20.4M & 38.57 & 0.9627 & 34.67 & 0.9261 & 32.58 & 0.9048 & 34.15 & 0.9446 & 40.28 & 0.9806 \\
    CAT-A~\cite{chen2022cross}      & $\times$2 & 16.5M & 38.51 & 0.9626 & 34.78 & 0.9265 & 32.59 & 0.9047 & 34.26 & 0.9440 & 40.10 & 0.9805 \\
    ART~\cite{zhang2023accurate}    & $\times$2 & 16.4M & 38.56 & 0.9629 & 34.59 & 0.9267 & 32.58 & 0.9048 & 34.30 & 0.9452 & 40.24 & 0.9808 \\
    HAT~\cite{chen2023activating}   & $\times$2 & 20.6M & 38.63 & 0.9630 & 34.86 & 0.9274 & \sotab{32.62} & \sotab{0.9053} & 34.45 & 0.9466 & 40.26 & 0.9809 \\

    MambaIRv2~\cite{guo2024mambairv2}& $\times$2 & 22.9M & \sotab{38.65} & \sotab{0.9632} & \sotab{34.89} & \sotab{0.9275} & \sotab{32.62} & \sotab{0.9053} & \sotab{34.49} & \sotab{0.9468} & \sotab{40.42} & \sotaa{0.9810} \\
    \rowcolor{Gray}
    \textbf{ATD} (ours)              & $\times$2 & 21.8M & \sotaa{38.68} & \sotaa{0.9633} & \sotaa{34.92} & \sotaa{0.9278} & \sotaa{32.67} & \sotaa{0.9059} & \sotaa{34.78} & \sotaa{0.9486} & \sotaa{40.51} & \sotaa{0.9814} \\

    % $\times$3 %%%%%%%%%%%%%%%%%%%%%%%%%%%%%%%%%%%%%%%%%%%%%%%%%%%
    \midrule

    EDSR~\cite{lim2017edsr}         & $\times$3 & 43.0M & 34.65 & 0.9280 & 30.52 & 0.8462 & 29.25 & 0.8093 & 28.80 & 0.8653 & 34.17 & 0.9476 \\
    RCAN~\cite{zhang2018rcan}       & $\times$3 & 15.6M & 34.74 & 0.9299 & 30.65 & 0.8482 & 29.32 & 0.8111 & 29.09 & 0.8702 & 34.44 & 0.9499 \\
    SAN~\cite{Dai_2020_san}         & $\times$3 & 15.9M & 34.75 & 0.9300 & 30.59 & 0.8476 & 29.33 & 0.8112 & 28.93 & 0.8671 & 34.30 & 0.9494 \\
    HAN~\cite{Niu_2020_han}         & $\times$3 & 64.2M & 34.75 & 0.9299 & 30.67 & 0.8483 & 29.32 & 0.8110 & 29.10 & 0.8705 & 34.48 & 0.9500 \\
    % IPT~\cite{Chen_2020_ipt}        & $\times$3 & 116M & 34.81 & -      & 30.85 & -      & 29.38 & -      & 29.49 & -      & -     & -      \\
    SwinIR~\cite{liang2021swinir}   & $\times$3 & 11.9M & 34.97 & 0.9318 & 30.93 & 0.8534 & 29.46 & 0.8145 & 29.75 & 0.8826 & 35.12 & 0.9537 \\
    % EDT~\cite{li2021efficient}      & $\times$3 & 11.6M & 34.97 & 0.9316 & 30.89 & 0.8527 & 29.44 & 0.8142 & 29.72 & 0.8814 & 35.13 & 0.9534 \\
    % MambaIR~\cite{guo2024mambair}   & $\times$3 & 20.4M & 35.08 & 0.923 & 30.99 & 0.8536 & 29.51 & 0.8157 & 29.93 & 0.8841 & 35.43 & 0.9546 \\
    CAT-A~\cite{chen2022cross}      & $\times$3 & 16.6M & 35.06 & 0.9326 & 31.04 & 0.8538 & 29.52 & 0.8160 & 30.12 & 0.8862 & 35.38 & 0.9546 \\
    ART~\cite{zhang2023accurate}    & $\times$3 & 16.6M & \sotab{35.07} & 0.9325 & 31.02 & 0.8541 & 29.51 & 0.8159 & 30.10 & 0.8871 & 35.39 & 0.9548 \\
    HAT~\cite{chen2023activating}   & $\times$3 & 20.8M & 35.07 & 0.9329 & 31.08 & 0.8555 & 29.54 & 0.8167 & 30.23 & 0.8896 & 35.53 & 0.9552 \\
    MambaIRv2~\cite{guo2024mambairv2}& $\times$3 & 23.1M & \sotaa{35.18} & \sotab{0.9334} & \sotaa{31.12} & \sotab{0.8557} & \sotab{29.55} & \sotab{0.8169} & \sotab{30.28} & \sotab{0.8905} & \sotab{35.61} & \sotab{0.9556} \\

    \rowcolor{Gray}
    \textbf{ATD} (ours)             & $\times$3 & 22.0M & \sotaa{35.18} & \sotaa{0.9335} & \sotab{31.11} & \sotaa{0.8559} & \sotaa{29.58} & \sotaa{0.8178} & \sotaa{30.55} & \sotaa{0.8931} & \sotaa{35.74} & \sotaa{0.9563} \\

    % $\times$4 %%%%%%%%%%%%%%%%%%%%%%%%%%%%%%%%%%%%%%%%%%%%%%%%%%%
    \midrule
    EDSR~\cite{lim2017edsr}         & $\times$4 & 43.0M & 32.46 & 0.8968 & 28.80 & 0.7876 & 27.71 & 0.7420 & 26.64 & 0.8033 & 31.02 & 0.9148 \\
    RCAN~\cite{zhang2018rcan}       & $\times$4 & 15.6M & 32.63 & 0.9002 & 28.87 & 0.7889 & 27.77 & 0.7436 & 26.82 & 0.8087 & 31.22 & 0.9173 \\
    SAN~\cite{Dai_2020_san}         & $\times$4 & 15.9M & 32.64 & 0.9003 & 28.92 & 0.7888 & 27.78 & 0.7436 & 26.79 & 0.8068 & 31.18 & 0.9169 \\
    HAN~\cite{Niu_2020_han}         & $\times$4 & 64.2M & 32.64 & 0.9002 & 28.90 & 0.7890 & 27.80 & 0.7442 & 26.85 & 0.8094 & 31.42 & 0.9177 \\
    % CSNLN~\cite{Mei2020image}       & $\times$4 & 7.16M & 32.68 & 0.9004 & 28.95 & 0.7888 & 27.80 & 0.7439 & 27.22 & 0.8168 & 31.43 & 0.9201 \\
    % IPT~\cite{Chen_2020_ipt}        & $\times$4 & 116M  & 32.64 & -      & 29.01 & -      & 27.82 & -      & 27.26 & -      & -     & -      \\   
    SwinIR~\cite{liang2021swinir}   & $\times$4 & 11.9M & 32.92 & 0.9044 & 29.09 & 0.7950 & 27.92 & 0.7489 & 27.45 & 0.8254 & 32.03 & 0.9260 \\
    % EDT~\cite{li2021efficient}      & $\times$4 & 11.6M & 32.82 & 0.9031 & 29.09 & 0.7939 & 27.91 & 0.7483 & 27.46 & 0.8246 & 32.05 & 0.9254 \\
    CAT-A~\cite{chen2022cross}      & $\times$4 & 16.6M & 33.08 & 0.9052 & 29.18 & 0.7960 & 27.99 & 0.7510 & 27.89 & 0.8339 & 32.39 & 0.9285 \\
    ART~\cite{zhang2023accurate}    & $\times$4 & 16.6M & 33.04 & 0.9051 & 29.16 & 0.7958 & 27.97 & 0.7510 & 27.77 & 0.8321 & 32.31 & 0.9283 \\
    HAT~\cite{chen2023activating}   & $\times$4 & 20.8M & 33.04 & 0.9056 & \sotab{29.23} & 0.7973 & \sotab{28.00} & \sotab{0.7517} & \sotab{27.97} & \sotab{0.8368} & 32.48 & 0.9292 \\
    MambaIRv2~\cite{guo2024mambairv2}& $\times$4 & 23.1M & \sotab{33.14} & \sotab{0.9057} & \sotab{29.23} & \sotab{0.7975} & \sotab{28.00} & 0.7511 & 27.89 & 0.8344 & \sotab{32.57} & \sotab{0.9295} \\
    
    \rowcolor{Gray}
    \textbf{ATD} (ours)             & $\times$4 & 22.0M & \sotaa{33.19} & \sotaa{0.9068} & \sotaa{29.30} & \sotaa{0.7986} & \sotaa{28.04} & \sotaa{0.7530} & \sotaa{28.24} & \sotaa{0.8420} & \sotaa{32.71} & \sotaa{0.9311} \\
    \bottomrule
  \end{tabular}
  \end{center}
\vspace{-2mm}
\end{table*}

\begin{table*}[t]
\scriptsize
% \footnotesize
\setlength{\tabcolsep}{8.3pt}

\caption{Quantitative comparison (PSNR/SSIM) with state-of-the-art methods on \textbf{lightweight image SR} task. Best and second best results are marked with \sotaa{bold} and \sotab{underline}. Experimental details can be found in Sec.~\ref{sec:srdetails}. }
\vspace{-2mm}
\label{tab: results lightweight sr}
  % \centering
  % \begin{tabular}{l|l}
  \begin{center}
      
  \begin{tabular}{l|c|c|cc|cc|cc|cc|cc}
    \toprule
    % \hline
    % \cmidrule(r){1-4}
    \multirow{2}{*}{\textbf{Method}} & \multirow{2}{*}{\textbf{Scale}} & \multirow{2}{*}{\textbf{Params}} & \multicolumn{2}{c|}{\textbf{Set5}~\cite{Bevilacqua2012set5}} & \multicolumn{2}{c|}{\textbf{Set14}~\cite{zeyde2012single}} & \multicolumn{2}{c|}{\textbf{BSD100}~\cite{MartinFTM01}} & \multicolumn{2}{c|}{\textbf{Urban100}~\cite{Huang_2015_Urban100}} & \multicolumn{2}{c}{\textbf{Manga109}~\cite{Matsui_2016_Manga109}} \\

    % $\times$2 %%%%%%%%%%%%%%%%%%%%%%%%%%%%%%%%%%%%%%%%%%%%%%%%%%%
    & & & PSNR & SSIM & PSNR & SSIM & PSNR & SSIM & PSNR & SSIM & PSNR & SSIM   \\
    \midrule

    CARN~\cite{Ahn_2018_carn}               & $\times$2 & 1,592K & 37.76 & 0.9590 & 33.52 & 0.9166 & 32.09 & 0.8978 & 31.92 & 0.9256 & 38.36 & 0.9765 \\
    IMDN~\cite{Hui_2019_imdn}               & $\times$2 & 694K   & 38.00 & 0.9605 & 33.63 & 0.9177 & 32.19 & 0.8996 & 32.17 & 0.9283 & 38.88 & 0.9774 \\
    LAPAR-A~\cite{Li_2020_lapar}            & $\times$2 & 548K   & 38.01 & 0.9605 & 33.62 & 0.9183 & 32.19 & 0.8999 & 32.10 & 0.9283 & 38.67 & 0.9772 \\
    LatticeNet~\cite{Luo_2020_latticenet}   & $\times$2 & 756K   & 38.15 & 0.9610 & 33.78 & 0.9193 & 32.25 & 0.9005 & 32.43 & 0.9302 & -     & -      \\
    SwinIR-light~\cite{liang2021swinir}     & $\times$2 & 910K   & 38.14 & 0.9611 & 33.86 & 0.9206 & 32.31 & 0.9012 & 32.76 & 0.9340 & 39.12 & 0.9783 \\
    % MambaIR-light~\cite{guo2024mambair}     & $\times$2 & 905K   & 38.13 & 0.9610 & 33.95 & 0.9208 & 32.31 & 0.9013 & 32.85 & 0.9349 & 39.20 & 0.9782 \\
    ELAN~\cite{zhang2022efficient}          & $\times$2 & 582K   & 38.17 & 0.9611 & 33.94 & 0.9207 & 32.30 & 0.9012 & 32.76 & 0.9340 & 39.11 & 0.9782 \\
    SwinIR-NG~\cite{Choi_2022_swinirng}     & $\times$2 & 1181K  & 38.17 & 0.9612 & 33.94 & 0.9205 & 32.31 & 0.9013 & 32.78 & 0.9340 & 39.20 & 0.9781 \\
    SRFormer-light~\cite{zhou2023srformer}  & $\times$2 & 853K  & 38.23 & 0.9615 & 33.94 & 0.9209 & 32.36 & 0.9013 & 32.91 & 0.9353 & 39.28 & 0.9785 \\
    OmniSR~\cite{omni_sr}                   & $\times$2 & 772K   & 38.22 & 0.9613 & 33.98 & 0.9210 & \sotab{32.36} & \sotab{0.9020} & 33.05 & 0.9363 & 39.28 & 0.9784 \\
    MambaIRv2-light~\cite{guo2024mambairv2} & $\times$2 & 774K   & \sotab{38.26} & \sotab{0.9615} & \sotab{34.09} & \sotab{0.9221} & \sotab{32.36} & 0.9019 & \sotab{33.26} & \sotab{0.9378} & \sotab{39.35} & \sotab{0.9785} \\
    
    \rowcolor{Gray}
    \textbf{ATD-light} (Ours)               & $\times$2 & 761K   & \sotaa{38.33} & \sotaa{0.9618} & \sotaa{34.11} & \sotaa{0.9224} & \sotaa{32.40} & \sotaa{0.9025} & \sotaa{33.36} & \sotaa{0.9387} & \sotaa{39.54} & \sotaa{0.9788} \\

    % $\times$3 %%%%%%%%%%%%%%%%%%%%%%%%%%%%%%%%%%%%%%%%%%%%%%%%%%%
    \midrule

    CARN~\cite{Ahn_2018_carn}               & $\times$3 & 1,592K & 34.29 & 0.9255 & 30.29 & 0.8407 & 29.06 & 0.8034 & 28.06 & 0.8493 & 33.50 & 0.9440 \\
    IMDN~\cite{Hui_2019_imdn}               & $\times$3 & 703K   & 34.36 & 0.9270 & 30.32 & 0.8417 & 29.09 & 0.8046 & 28.17 & 0.8519 & 33.61 & 0.9445 \\
    LAPAR-A~\cite{Li_2020_lapar}            & $\times$3 & 544K   & 34.36 & 0.9267 & 30.34 & 0.8421 & 29.11 & 0.8054 & 28.15 & 0.8523 & 33.51 & 0.9441 \\
    LatticeNet~\cite{Luo_2020_latticenet}   & $\times$3 & 765K   & 34.53 & 0.9281 & 30.39 & 0.8424 & 29.15 & 0.8059 & 28.33 & 0.8538 & -     & -      \\
    SwinIR-light~\cite{liang2021swinir}     & $\times$3 & 918K   & 34.62 & 0.9289 & 30.54 & 0.8463 & 29.20 & 0.8082 & 28.66 & 0.8624 & 33.98 & 0.9478 \\
    ELAN~\cite{zhang2022efficient}          & $\times$3 & 590K   & 34.61 & 0.9288 & 30.55 & 0.8463 & 29.21 & 0.8081 & 28.69 & 0.8624 & 34.00 & 0.9478 \\
    SwinIR-NG~\cite{Choi_2022_swinirng}     & $\times$3 & 1190K  & 34.64 & 0.9293 & 30.58 & 0.8471 & 29.24 & 0.8090 & 28.75 & 0.8639 & 34.22 & 0.9488 \\
    OmniSR~\cite{omni_sr}                   & $\times$3 & 780K   & 34.70 & 0.9294 & 30.57 & 0.8469 & 29.28 & 0.8094 & 28.84 & 0.8656 & 34.22 & 0.9487 \\
    MambaIRv2-light~\cite{guo2024mambairv2} & $\times$3 & 781K   & \sotab{34.71} & \sotab{0.9298} & \sotab{30.68} & \sotab{0.8483} & \sotab{29.26} & \sotab{0.8098} & \sotab{29.01} & \sotab{0.8689} & \sotab{34.41} & \sotab{0.9497} \\
    
    \rowcolor{Gray}
    \textbf{ATD-light} (ours)               & $\times$3 & 767K   & \sotaa{34.75} & \sotaa{0.9302} & \sotaa{30.71} & \sotaa{0.8489} & \sotaa{29.33} & \sotaa{0.8113} & \sotaa{29.26} & \sotaa{0.8726} & \sotaa{34.61} & \sotaa{0.9508} \\

    % $\times$4 %%%%%%%%%%%%%%%%%%%%%%%%%%%%%%%%%%%%%%%%%%%%%%%%%%%
    \midrule
    CARN~\cite{Ahn_2018_carn}               & $\times$4 & 1,592K & 32.13 & 0.8937 & 28.60 & 0.7806 & 27.58 & 0.7349 & 26.07 & 0.7837 & 30.47 & 0.9084 \\
    IMDN~\cite{Hui_2019_imdn}               & $\times$4 & 715K   & 32.21 & 0.8948 & 28.58 & 0.7811 & 27.56 & 0.7353 & 26.04 & 0.7838 & 30.45 & 0.9075 \\
    LAPAR-A~\cite{Li_2020_lapar}            & $\times$4 & 659K   & 32.15 & 0.8944 & 28.61 & 0.7818 & 27.61 & 0.7366 & 26.14 & 0.7871 & 30.42 & 0.9074 \\
    LatticeNet~\cite{Luo_2020_latticenet}   & $\times$4 & 777K   & 32.30 & 0.8962 & 28.68 & 0.7830 & 27.62 & 0.7367 & 26.25 & 0.7873 & -     & -      \\
    SwinIR-light~\cite{liang2021swinir}     & $\times$4 & 930K   & 32.44 & 0.8976 & 28.77 & 0.7858 & 27.69 & 0.7406 & 26.47 & 0.7980 & 30.92 & 0.9151 \\
    ELAN~\cite{zhang2022efficient}          & $\times$4 & 582K   & 32.43 & 0.8975 & 28.78 & 0.7858 & 27.69 & 0.7406 & 26.54 & 0.7982 & 30.92 & 0.9150 \\
    SwinIR-NG~\cite{Choi_2022_swinirng}     & $\times$4 & 1201K  & 32.44 & 0.8980 & 28.83 & 0.7870 & 27.73 & 0.7418 & 26.61 & 0.8010 & 31.09 & 0.9161 \\
    OmniSR~\cite{omni_sr}                   & $\times$4 & 792K   & 32.49 & 0.8988 & 28.78 & 0.7859 & 27.71 & 0.7415 & 26.65 & 0.8018 & 31.02 & 0.9151 \\
    MambaIRv2-light~\cite{guo2024mambairv2} & $\times$4 & 790K   & \sotab{32.51} & \sotab{0.8992} & \sotab{28.84} & \sotab{0.7878} & \sotab{27.75} & \sotab{0.7426} & \sotab{26.82} & \sotab{0.8079} & \sotab{31.24} & \sotab{0.9182} \\

    \rowcolor{Gray}
    \textbf{ATD-light} (Ours)      & $\times$4 & 776K   & \sotaa{32.59} & \sotaa{0.8997} & \sotaa{28.96} & \sotaa{0.7895} & \sotaa{27.79} & \sotaa{0.7440} & \sotaa{27.04} & \sotaa{0.8127} & \sotaa{31.52} & \sotaa{0.9203} \\
    \bottomrule
    % \hline
  \end{tabular}
  \end{center}
\vspace{-2mm}
\end{table*}

\begin{figure*}[h]
    \centering
    \includegraphics[width=\linewidth]{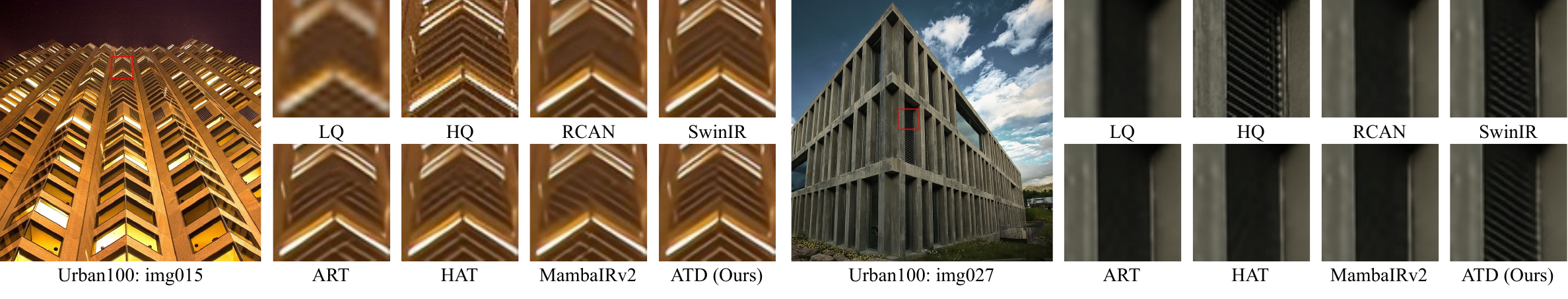}
    \includegraphics[width=\linewidth]{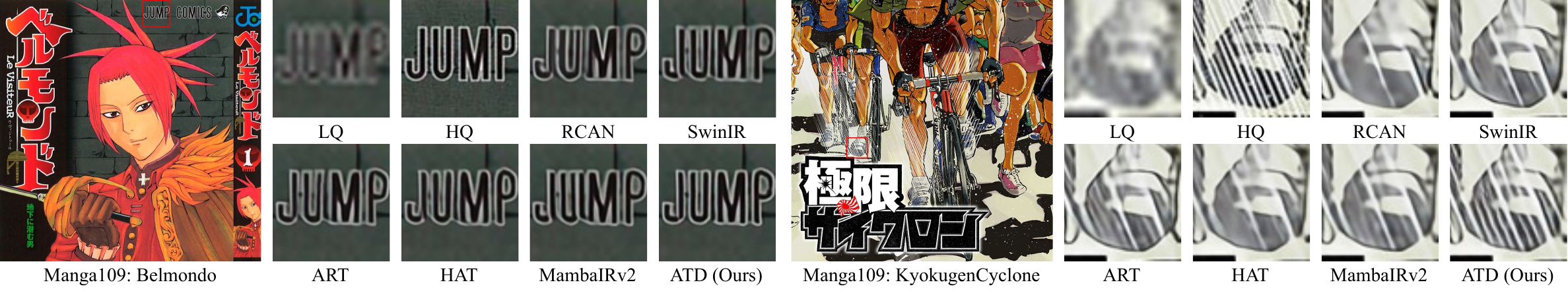}
    \caption{Qualitative comparison on $\times 4$ image super-resolution.}
    \label{fig:comp_sr_x4}
\end{figure*}

\noindent
\textbf{Training Settings.}
Following existing works~\cite{liang2021swinir, chen2023activating}, we apply DIV2K~\cite{timofte2017div2k} as the training dataset for ATD-light, and augment it with Flickr2K~\cite{lim2017edsr} to form DF2K for training the full ATD model.
Both models are optimized using AdamW~\cite{loshchilov2018decoupled} with $\beta = [0.9, 0.9]$.
For ATD-light, the $\times2$ model is trained for 500K iterations with batch size of 64 and patch size of 64$\times$64. 
The initial learning rate is set to $5 \times 10^{-4}$ and halved at [250K, 400K, 450K, 480K] iterations.
The $\times 3$ and $\times 4$ models are finetuned from the pretrained $\times 2$ model for 250K iterations each.
For the full ATD model, we first pretrain a $\times 2$ model for 300K iterations with batch size of 48 and patch size of 64$\times$64. 
Finetuning is then performed for all scaling factors on the pretrained model using a larger patch size of $96 \times 96$ for 120K iterations.
The learning rate is initialized as $2\times10^{-4}$ and halved at 80K and 100K iterations.
The training process is conducted on 4 NVIDIA RTX 4090 GPUs, with both pretraining and finetuning stages taking approximately 4 days each.

\subsection{Quantitative results}
In Tab.~\ref{tab: results classical sr}, we present the qualitative comparison on classical image SR task, evaluating ATD against state-of-the-art methods: EDSR~\cite{lim2017edsr}, RCAN~\cite{zhang2018rcan}, SAN~\cite{Dai_2020_san}, HAN~\cite{Niu_2020_han}, IPT~\cite{Chen_2020_ipt}, SwinIR~\cite{liang2021swinir}, MambaIR~\cite{guo2024mambair}, CAT~\cite{chen2022cross}, ART~\cite{zhang2023accurate}, HAT~\cite{chen2023activating}, and MambaIRv2~\cite{guo2024mambairv2}.
ATD consistently outperforms these methods across five benchmark datasets.
Notably, with a comparable model size, ATD surpasses HAT by  0.29-0.40 dB on the Urban100 and Manga109 datasets.
Compared with MambaIRv2, which also models global dependencies, ATD achieves performance gains of 0.27dB-0.35dB on Urban100.
It highlights the superior modeling capability of the proposed ACMSA module in leveraging the repetitive structures in Urban100.
The computational overhead results are provided in Tab.~\ref{tab:comp_modelsize}, where we compare the FLOPs, inference time, and GPU memory footprint of ATD with state-of-the-art approaches.
It shows that ATD strikes a superior balance between performance and computational efficiency, particularly when compared to ART which also utilizes sparse attention.
Meanwhile, ATD requires approximately $30\%$ less GPU memory than HAT and achieves $25$-$50\%$ faster inference speed than MambaIRv2 while delivering performance improvements of up to 0.35 dB, justifying the modest $20\%$ increase in FLOPs.
In addition to the quantitative comparison, a detailed analysis of computational complexity is provided in the appendix.
In Tab.~\ref{tab: results lightweight sr}, we further compare ATD-light with state-of-the-art lightweight SR models: CARN~\cite{Ahn_2018_carn}, IMDN~\cite{Hui_2019_imdn}, LAPAR~\cite{Li_2020_lapar}, LatticeNet~\cite{Luo_2020_latticenet}, SwinIR-light~\cite{liang2021swinir}, MambaIR-light~\cite{guo2024mambair}, ELAN~\cite{zhang2022efficient}, SRFormer-light~\cite{zhou2023srformer}, OmniSR~\cite{omni_sr}, and MambaIRv2-light~\cite{guo2024mambairv2}.
ATD-light achieves the best performance among all compared lightweight models, with improvements of 0.10-0.28 dB on Urban100 and Manga109 datasets.
In particular, for SR factor of $\times 4$, ATD-light surpasses MambaIRv2-light by 0.22-0.28dB while maintaining a similar model size, demonstrating its strong capability in reconstructing heavily degraded images.
In addition to synthetic datasets, we also train an ATD model for real-world image super-resolution.
Due to the page limit, the results are included in the appendix.

\subsection{Visual comparisons}
In Fig.~\ref{fig:comp_sr_x4} we present qualitative comparisons between ATD and several recent SOTA image SR methods. 
In challenging scenarios where high-frequency details are severely degraded, most existing methods fail to reconstruct the accurate image structures. 
For instance, in \textit{img015} of Urban100 and \textit{Belmondo} of Manga109, most methods produce blurry and distorted edges, with some even producing lines in incorrect directions.
Similarly, in images \textit{img027} of Urban100 and \textit{KyokugenCyclone} of Manga109, where line structures are almost missing in the low-quality input, these methods struggle to recover plausible details.
In contrast, ATD successfully restores sharper and structurally better content across these examples, highlighting its superiority in preserving fine-grained details and geometric fidelity in super-resolution tasks.

\begin{table*}[t]
\setlength{\tabcolsep}{3.5pt}
    \caption{Comparisons on computational overhead. We use single RTX4090 GPU to test the runtime and GPU memory footprint for various LQ input sizes. ``OOM" denotes out-of-memory.}
    \label{tab:comp_modelsize}
    \vspace{-2mm}
    \centering
    \begin{tabular}{l|ccc|ccc|ccc|cc|cc}
    \toprule
        \multirow{2}{*}{\textbf{Model}} & \multicolumn{3}{c|}{\textbf{$128\times128$}} & \multicolumn{3}{c|}{\textbf{$256\times256$}} & \multicolumn{3}{c|}{\textbf{$512\times512$}} & \multicolumn{2}{c|}{\textbf{Urban100}~\cite{Huang_2015_Urban100}} & \multicolumn{2}{c}{\textbf{Manga109}~\cite{Matsui_2016_Manga109}} \\
        & FLOPs & Runtime & Memory & FLOPs & Runtime & Memory & FLOPs & Runtime & Memory & PSNR & SSIM & PSNR & SSIM \\
        \midrule
        ART~\cite{zhang2023accurate}        & 0.39T & 104ms & 1565M & 2.85T & 1237ms & 15037M & 32.3T & - & OOM & 27.77 & 0.8321 & 32.31 & 0.9283 \\
        HAT~\cite{chen2023activating}       & 0.41T & 105ms & 1443M & 1.65T & 450ms & 4057M & 6.60T & 1950ms & 12855M & \sotab{27.97} & \sotab{0.8368} & 32.48 & 0.9292 \\
        MambaIRv2~\cite{guo2024mambairv2}   & 0.45T & 175ms & 1057M & 1.78T & 821ms & 2371M & 7.13T & 3830ms & 7727M & 27.89 & 0.8344 & \sotab{32.57} & \sotab{0.9295} \\
        ATD (Ours)                          & 0.51T & 137ms & 1025M & 2.03T & 612ms & 2385M & 8.11T & 2530ms & 8681M & \textbf{28.24} & \textbf{0.8420} & \textbf{32.71} & \textbf{0.9311} \\

    \bottomrule
    \end{tabular}
\vspace{-2mm}
\end{table*}

\subsection{Ablation Studies}
In this subsection, we conduct a series of ablation studies to validate the efficacy of the proposed components.
We conduct these experiments on $\times 4$ ATD-light, trained on DIV2K~\cite{timofte2017div2k} dataset for 250K iterations.
We apply Urban100~\cite{Huang_2015_Urban100} and Manga109~\cite{Matsui_2016_Manga109} datasets as evaluation benchmarks.
\begin{table}[t]
\setlength{\tabcolsep}{10pt}
    \caption{Ablation studies on proposed key components TDCA, ACMSA, and CFFN in metric of PSNR.}
    \label{tab:ablation_comp}
\vspace{-2mm}
    \centering
    \begin{tabular}{l|cc|cc}
    \toprule
        \multirow{2}{*}{\textbf{Model}} & \multicolumn{2}{c|}{\textbf{Urban100}~\cite{Huang_2015_Urban100}} & \multicolumn{2}{c}{\textbf{Manga109}~\cite{Matsui_2016_Manga109}} \\
        & PSNR & SSIM & PSNR & SSIM \\
        \midrule
        baseline & 26.40 & 0.7956 & 30.84 & 0.9131 \\
        +TDCA    & 26.46 & 0.7971 & 30.92 & 0.9140 \\
        +ACMSA   & 26.70 & 0.8023 & 31.19 & 0.9165 \\
        +CFFN    & \sotaa{26.76} & \sotaa{0.8039} & \sotaa{31.21} & \sotaa{0.9169} \\

    \bottomrule
    \end{tabular}
\vspace{-2mm}
\end{table}

\noindent
\textbf{Effects of the proposed TDCA, ACMSA, and CFFN.}
The core components of ATD include Token Dictionary Cross-Attention (TDCA), Adaptive Category-based MSA (ACMSA), and Category-aware FFN (CFFN). 
In this ablation, we construct four model variants to evaluate the efficacy of each component, with results presented in Tab.~\ref{tab:ablation_comp}. 
We adopt \cite{liang2021swinir} as the baseline and progressively incorporate the proposed modules.
Firstly, TDCA fuses the external information (e.g. typical image structures) and input features through a cross-attention mechanism, yielding improvements of 0.06 dB on Urban100 and 0.08 dB on Manga109.
Notably, TDCA not only contributes directly to performance gains, but also provides essential category information for ACMSA.
By leveraging the category information embedded in the attention maps of TDCA, ACMSA facilitates global dependencies modeling and achieves significant performance gains of 0.24-0.27 dB.
Finally, incorporating CFFN to further enhance category-aware feature fusion, the complete ATD model achieves the best results of 26.76 dB and 31.21 dB on the evaluation datasets, demonstrating the effectiveness of proposed architectures.
We also provide a detailed comparison between the proposed category-based self-attention with the window-based baseline, including analysis on convergence properties.
These results can be found in the appendix, which show that the proposed method delivers superior results with lower computataional complexity.

\begin{table}[t]
\setlength{\tabcolsep}{10pt}
    \caption{Ablation studies on improved scaling factor $\tau'$ and choice of dictionary size $M$.}
    \label{tab:ablation_dictsize}
\vspace{-2mm}
    \centering
    \begin{tabular}{cc|cc|cc}
    \toprule
        \multirow{2}{*}{$\tau'$} & \multirow{2}{*}{$M$} & \multicolumn{2}{c|}{\textbf{Urban100}~\cite{Huang_2015_Urban100}} & \multicolumn{2}{c}{\textbf{Manga109}~\cite{Matsui_2016_Manga109}} \\
        & & PSNR & SSIM & PSNR & SSIM \\
        \midrule
                & 32  & 26.70 & 0.8019 & 31.15 & 0.9162 \\
                & 64  & 26.73 & 0.8029 & 31.15 & 0.9160 \\
                & 128 & 26.70 & 0.8026 & 31.19 & 0.9167 \\
        % & 256 & & & & & & \\
        \midrule
        $\surd$ & 32  & 26.70 & 0.8022 & 31.17 & 0.9163 \\
        $\surd$ & 64  & 26.72 & 0.8027 & 31.19 & 0.9165 \\
        $\surd$ & 128 & \sotaa{26.76} & \sotaa{0.8039} & \sotaa{31.21} & \sotaa{0.9169} \\
        % $\surd$ & 256 & & & & & & \\

    \bottomrule
    \end{tabular}
\vspace{-2mm}
\end{table}

\begin{table}[t]
\setlength{\tabcolsep}{10pt}
    \caption{Ablation studies on choice of sub-category size $n_s$. $n_s=0$ represents removing the ACMSA branch.}
    \label{tab:ablation_subcatsize}
\vspace{-2mm}
    \centering
    \begin{tabular}{c|cc|cc}
    \toprule
        \multirow{2}{*}{$n_s$} & \multicolumn{2}{c|}{\textbf{Urban100}~\cite{Huang_2015_Urban100}} & \multicolumn{2}{c}{\textbf{Manga109}~\cite{Matsui_2016_Manga109}} \\
        & PSNR & SSIM & PSNR & SSIM \\
        \midrule
        0   & 26.46 & 0.7971 & 30.92 & 0.9140 \\
        32  & 26.66 & 0.8015 & 31.13 & 0.9160 \\
        64  & 26.73 & 0.8028 & 31.19 & 0.9170 \\
        128 & \textbf{26.76} & \textbf{0.8039} & 31.21 & 0.9169 \\
        256 & 26.75 & 0.8036 & \textbf{31.23} & \textbf{0.9172} \\

    \bottomrule
    \end{tabular}
% \vspace{-2mm}
\end{table}

\noindent
\textbf{Effects of dictionary size $M$ and reparameterized scaling factor $\tau'$.}
In Tab.~\ref{tab:ablation_dictsize}, we report the performance of models with varying dictionary sizes $M=[32,64,128,256]$, under two types of scaling factors ($\tau$ and $\tau'$).
With the original scaling factor $\tau$, the model achieves its best performance when $M=64$, followed by a performance decline as the dictionary size increases.
This degradation arises because, in an excessively large representation space, the softmax operation tends to suppress the weights of highly relevant dictionary entries.
As a result, the model struggles to effectively distinguish between highly and weakly correlated dictionary entries, leading to suboptimal performance.
In contrast, the improved scaling factor $\tau'$ adaptively enlarges the gap between attention weights according to the dictionary size, leading to continuous performance improvements as $M$ increases.
To balance performance and computational overhead, we set $M=128, 512, 128$ for ATD-light, ATD, and ATD-U, respectively.

\begin{table*}[t]
% \vspace{-4mm}
% \setlength{\belowcaptionskip}{0.2cm}
% \tiny
\scriptsize
% \footnotesize
\setlength{\tabcolsep}{1.4pt}

\vspace{2mm}
\caption{Quantitative comparison (PSNR/SSIM) with state-of-the-art methods on \textbf{color image denoising} task. Best and second best results are marked with \sotaa{bold} and \sotab{underline}. Experimental details can be found in Sec.~\ref{sec:dnjpegdetails}. }
\label{tab: results color image dn}
\vspace{-3mm}
\begin{center}
\begin{tabular}{l|c|ccccccccccc}

    \toprule

    \textbf{Dataset} & $\sigma$ & BM3D~\cite{dabov2007image} & DnCNN~\cite{zhang2017beyond} & IRCNN~\cite{zhang2017learning} & FFDNet~\cite{zhang2018ffdnet} & RNAN~\cite{zhang2019rnan} & DRUNet~\cite{zhang2021plug} & SwinIR~\cite{liang2021swinir} & Restormer~\cite{Zamir2021Restormer} & SCUNet~\cite{zhang2023practical} & Xformer~\cite{zhang2023xformer} & \textbf{ATD-U} (ours) \\

    \midrule
    
    \multirow{3}{*}{\textbf{CBSD68}~\cite{MartinFTM01}}
    & 15 & 33.52 & 33.90 & 33.86 & 33.87 & -     & 34.30 & 34.42 & 34.40 & 33.40 & \sotab{34.43} & \sotaa{34.47} \\
    & 25 & 30.71 & 31.24 & 31.16 & 31.21 & -     & 31.69 & 31.78 & 31.79 & 31.79 & \sotab{31.82} & \sotaa{31.85} \\
    & 50 & 27.38 & 27.95 & 27.86 & 27.96 & 28.27 & 28.51 & 28.56 & 28.60 & 28.61 & \sotab{28.63} & \sotaa{28.66} \\

    \midrule
    
    \multirow{3}{*}{\textbf{Kodak24}~\cite{kodak1999}}
    & 15 & 34.28 & 34.60 & 34.69 & 34.63 & -     & 35.31 & 35.34 & 35.35 & 35.34 & \sotab{35.39} & \sotaa{35.57} \\
    & 25 & 32.15 & 32.14 & 32.18 & 32.13 &  -    & 32.89 & 32.89 & 32.93 & 32.92 & \sotab{32.99} & \sotaa{33.12} \\
    & 50 & 28.46 & 28.95 & 28.93 & 28.98 & 29.58 & 29.86 & 29.79 & 29.87 & 29.87 & \sotab{29.94} & \sotaa{30.12} \\
    
    \midrule
    
    \multirow{3}{*}{\textbf{McMaster}~\cite{zhang2011color}}
    & 15 & 34.06 & 33.45 & 34.58 & 34.66 & -     & 35.40 & 35.61 & 35.61 & 35.60 & \sotab{35.68} & \sotaa{35.72} \\
    & 25 & 31.66 & 31.52 & 32.18 & 32.35 & -     & 33.14 & 33.20 & 33.34 & 33.34 & \sotab{33.44} & \sotaa{33.47} \\
    & 50 & 28.51 & 28.62 & 28.91 & 29.18 & 29.72 & 30.08 & 30.22 & 30.30 & 30.29 & \sotab{30.38} & \sotaa{30.41} \\
    
    \midrule

    \multirow{3}{*}{\textbf{Urban100}~\cite{Huang_2015_Urban100}}
    & 15 & 33.93 & 32.98 & 33.78 & 33.83 & -     & 34.81 & 35.13 & 35.13 & 35.18 & \sotab{35.29} & \sotaa{35.37} \\
    & 25 & 31.36 & 30.81 & 31.20 & 31.40 & -     & 32.60 & 32.90 & 32.96 & 33.03 & \sotab{33.21} & \sotaa{33.28} \\
    & 50 & 27.93 & 27.59 & 27.70 & 28.05 & 29.08 & 29.61 & 29.82 & 30.02 & 30.14 & \sotab{30.36} & \sotaa{30.48} \\

    \bottomrule
  \end{tabular}
  \end{center}

% \vspace{-0.7cm}
  
\end{table*}

\begin{table*}[t]
% \vspace{-4mm}
% \setlength{\belowcaptionskip}{0.2cm}
% \tiny
\scriptsize
% \footnotesize
\setlength{\tabcolsep}{1.4pt}

\vspace{-3mm}

\caption{Quantitative comparison (PSNR/SSIM) with state-of-the-art methods on \textbf{grayscale image denoising} task. Best and second best results are marked with \sotaa{bold} and \sotab{underline}. Experimental details can be found in Sec.~\ref{sec:dnjpegdetails}. }
\label{tab: results grayscale image dn}
\vspace{-3mm}
  
\begin{center}
\begin{tabular}{l|c|ccccccccccc}

    \toprule

    \textbf{Dataset} & $\sigma$ & BM3D~\cite{dabov2007image} & DnCNN~\cite{zhang2017beyond} & IRCNN~\cite{zhang2017learning} & FFDNet~\cite{zhang2018ffdnet} & RNAN~\cite{zhang2019rnan} & DRUNet~\cite{zhang2021plug} & SwinIR~\cite{liang2021swinir} & Restormer~\cite{Zamir2021Restormer} & SCUNet~\cite{zhang2023practical} & Xformer~\cite{zhang2023xformer} & \textbf{ATD-U} (ours) \\

    \midrule
    
    \multirow{3}{*}{\textbf{Set12}~\cite{zhang2017beyond}}
    & 15 & 32.37 & 32.86 & 32.76 & 32.75 & -     & 33.25 & 33.36 & 33.42 & 33.43 & \sotab{33.46} & \sotaa{33.49} \\
    & 25 & 29.97 & 30.44 & 30.37 & 30.43 & -     & 30.94 & 31.01 & 31.08 & 31.09 & \sotaa{31.16} & \sotab{31.13} \\
    & 50 & 26.72 & 27.18 & 27.12 & 27.32 & 27.70 & 27.90 & 27.91 & 28.00 & 28.04 & \sotaa{28.10} & \sotaa{28.10} \\

    \midrule
    
    \multirow{3}{*}{\textbf{BSD68}~\cite{MartinFTM01}}
    & 15 & 31.08 & 31.73 & 31.63 & 31.63 & -     & 31.91 & 31.97 & 31.96 & \sotab{31.99} & 31.98 & \sotaa{32.02} \\
    & 25 & 28.57 & 29.23 & 29.15 & 29.19 &  -    & 29.48 & 29.50 & 29.52 & \sotab{29.55} & \sotab{29.55} & \sotaa{29.56} \\
    & 50 & 25.60 & 26.23 & 26.19 & 26.29 & 26.48 & 26.59 & 26.58 & 26.62 & \sotaa{26.67} & \sotab{26.65} & \sotab{26.65} \\
    
    \midrule

    \multirow{3}{*}{\textbf{Urban100}~\cite{Huang_2015_Urban100}}
    & 15 & 32.35 & 32.64 & 32.46 & 32.40 & -     & 33.44 & 33.70 & 33.79 & 33.88 & \sotab{33.98} & \sotaa{34.08} \\
    & 25 & 29.70 & 29.95 & 29.80 & 29.90 & -     & 31.11 & 31.30 & 31.46 & 31.58 & \sotaa{31.78} & \sotab{31.76} \\
    & 50 & 25.95 & 26.26 & 26.22 & 26.50 & 27.65 & 27.96 & 27.98 & 28.29 & 28.56 & \sotab{28.71} & \sotaa{28.80} \\

    \bottomrule
  \end{tabular}
  \end{center}

\vspace{-3mm}
  
\end{table*}

\noindent
\textbf{Effects of sub-category size $n_s$.}
In Tab.~\ref{tab:ablation_subcatsize}, we evaluate the impact of the key hyperparameter for ACMSA branch, i.e., the sub-category size $n_s$.
Analogous to the window size in the SWMSA~\cite{liu2021swin} branch, the sub-category size governs the granularity of token grouping.
Consistent with prior findings~\cite{chen2023activating, long2025progressive}, where a larger window size yields improved performance due to an expanded receptive field, increasing $n_s$ likewise leads to performance gains.
However, unlike SWMSA, the ACMSA branch is not constrained to local neighborhoods and is capable of modeling global dependencies.
As a result, ACMSA provides notable improvements even with a relatively small sub-category size.
To balance reconstruction quality and computational efficiency, we set $n_s = 128$ for ATD-light and ATD-U, and $n_s = 256$ for ATD.

\section{Experiments on Image Denoising and JPEG compression artifacts removal}
\label{sec:dnjpegexp}
\subsection{Implementation Details}
\label{sec:dnjpegdetails}
\noindent
\textbf{Network Architecture Details.}
For Gaussian image denoising and JPEG compression artifact removal tasks, we design ATD-U based on the U-Net architecture~\cite{Zamir2021Restormer}.
The numbers of Transformer layers in each encoder-decoder level and in the refinement block are set to [6, 8, 8, 6] and 6, respectively.
The number of channels and reduced channel dimension $d_r$ are set as [48, 96, 192, 384], 96, and [6, 12, 24, 24], 6, respectively.
Both the token dictionary size and sub-category size are set to 128.

\noindent
\textbf{Training Settings.}
We train the ATD-U models on the DFWB dataset, which comprises DIV2K~\cite{timofte2017div2k}, Flickr2K~\cite{lim2017edsr}, WED~\cite{ma2017waterloo}, and BSD500~\cite{amfm_pami2011}.
Muon~\cite{jordan2024muon} optimizer is employed with a weight decay of $1\times 10^{-2}$ to minimize the Charbonnier loss~\cite{charbonnier1994two} over 100K iterations.
The training patch and batch sizes are initially set to ($256^2$, 24), and are then changed to ($384^2$, 12) and ($512^2$, 6) at 60K and 80K iterations, respectively.
The learning rate is initially set to $1\times10^{-3}$ and halved at [50K, 70K, 90K] iterations.
The full training process takes approximately 4.5 days on 4 NVIDIA RTX4090 GPUs.
For grayscale image denoising, we use Set12~\cite{zhang2017beyond}, BSD68~\cite{MartinFTM01}, and Urban100~\cite{Huang_2015_Urban100} as evaluation datasets.
For color image denoising, CBSD~\cite{MartinFTM01}, Kodak24~\cite{kodak1999}, McMaster~\cite{zhang2011color}, and Urban100~\cite{Huang_2015_Urban100} are employed.
Meanwhile, the JPEG CAR benchmark includes Classic5~\cite{foi2007pointwise},  LIVE1~\cite{sheikh2006statistical} and Urban100~\cite{Huang_2015_Urban100}.

\subsection{Quantitative results}
\noindent
\textbf{Image denoising.}
Tab.~\ref{tab: results color image dn} and Tab.~\ref{tab: results grayscale image dn} present quantitative comparisons of the proposed ATD-U against state-of-the-art methods on both color and grayscale image denoising tasks, including BM3D~\cite{dabov2007image}, DnCNN~\cite{zhang2017beyond}, IRCNN~\cite{zhang2017learning}, FFDNet~\cite{zhang2018ffdnet}, RNAN~\cite{zhang2019rnan}, DRUNet~\cite{zhang2021plug}, SwinIR~\cite{liang2021swinir}, Restormer~\cite{Zamir2021Restormer}, SCUNet~\cite{zhang2023practical}, and Xformer~\cite{zhang2023xformer}.
For color image denoising, ATD-U consistently outperforms existing approaches across various benchmark datasets and noise levels, while maintaining a comparable model size (24.9M vs. 25.2M of Xformer).
In particular, it achieves greater improvements on datasets containing high-resolution images, with gains of up to 0.18 dB on Kodak24 and 0.12 dB on Urban100 across different noise levels.
These results underscore the effectiveness of our design in capturing long-range dependencies.
For grayscale image denoising, ATD-U also yields comparable or superior results across various benchmarks and noise levels.

\begin{figure*}
    \centering
    \includegraphics[width=\linewidth]{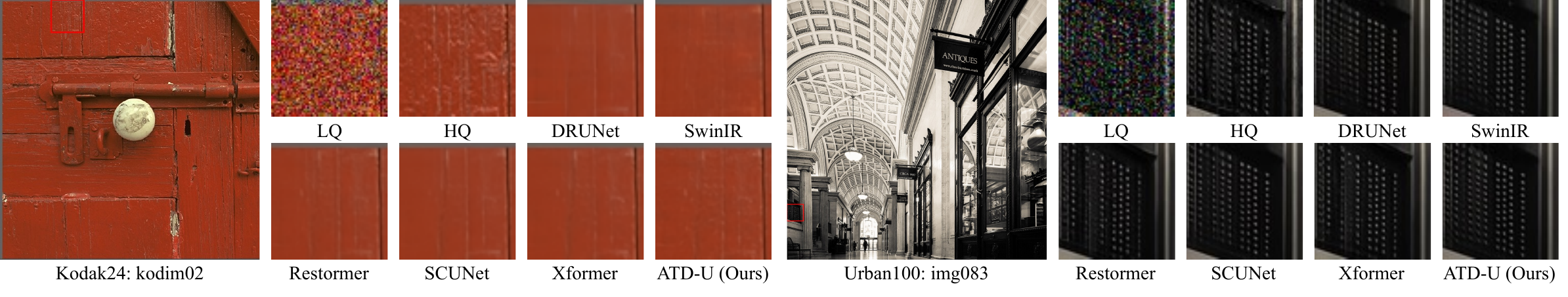}
    \caption{Qualitative comparison on color Gaussian image denoising with noise level $\sigma=50$.}
    \label{fig:comp_dn_color}
\end{figure*}
\begin{figure*}
    \centering
    \includegraphics[width=\linewidth]{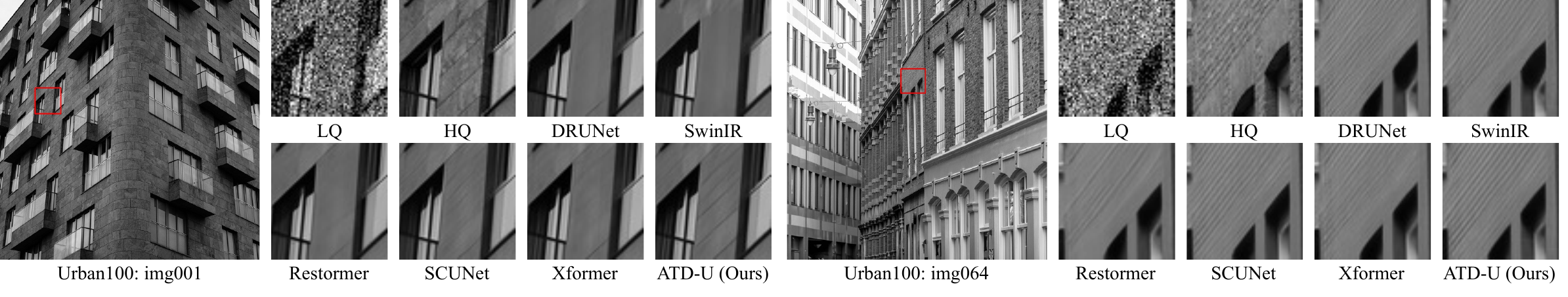}
    \caption{Qualitative comparison on grayscale Gaussian image denoising with noise level $\sigma=50$.}
    \label{fig:comp_dn_gray}
\end{figure*}

\noindent
\textbf{JPEG CAR.}
Tab.\ref{tab: results jpeg car} reports the quantitative results of ATD-U on the JPEG compression artifact removal (CAR) task, compared with state-of-the-art methods: ARCNN~\cite{yu2016deep}, DnCNN~\cite{zhang2017beyond}, RNAN~\cite{zhang2019rnan}, FBCNN~\cite{Jiang_2021_ICCV}, DRUNet~\cite{zhang2021plug}, SwinIR~\cite{liang2021swinir}, ART~\cite{zhang2023accurate}, and MambaIR~\cite{guo2024mambair}.
On Classic5, ATD-U surpasses ART and MambaIR by 0.05-0.07 dB in PSNR under various JPEG quality factors.
On Urban100, it achieves competitive performance with recent methods, and significantly outperforms the earlier transformer-based method SwinIR by 0.33dB at quality level $q=10$.
These gains demonstrate the capability of ATD-U in modeling global dependencies under severe compression, particularly for high-resolution images.

\subsection{Visual comparisons}
\begin{figure*}
    \centering
    \includegraphics[width=\linewidth]{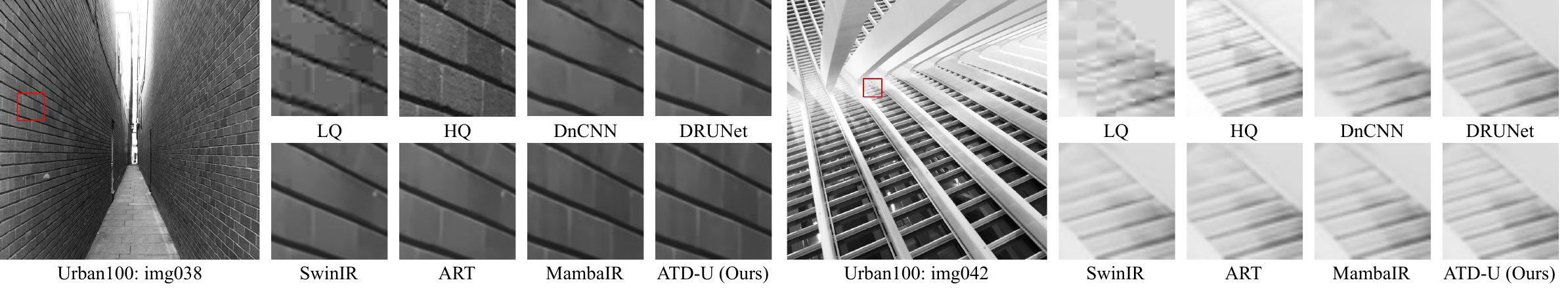}
    \caption{Qualitative comparison on JPEG CAR with quality factor $q=10$.}
    \label{fig:comp_jpegcar_gray}
\end{figure*}

In Fig.~\ref{fig:comp_dn_color}, \ref{fig:comp_dn_gray}, and \ref{fig:comp_jpegcar_gray}, we present the qualitative comparisons between ATD-U and existing state-of-the-art methods.
In these challenging tasks, noise and compression artifacts severely degrade fine textures and structures, making it difficult for existing methods to recover sharp and well-defined edges.
By contrast, ATD-U produces clearer and more visually appealing results, effectively restoring both irregular structures (e.g., \textit{kodim02} from Kodak24) and repetitive patterns (e.g., \textit{img064} from Urban100).
These qualitative results further validate the effectiveness of our global dependency modeling modules in mitigating noise and compression artifacts.

\begin{table*}[hbtp]
% \vspace{-4mm}
% \setlength{\belowcaptionskip}{0.2cm}
% \tiny
\scriptsize
% \footnotesize
\setlength{\tabcolsep}{3.6pt}

\vspace{2mm}
\caption{Quantitative comparison (PSNR/SSIM) with state-of-the-art methods on \textbf{JPEG compression artifacts removal} task. Best and second best results are marked with \sotaa{bold} and \sotab{underline}. Experimental details can be found in Sec.~\ref{sec:dnjpegdetails}. }
\label{tab: results jpeg car}
\vspace{-2mm}
  
\begin{center}
\begin{tabular}{l|c|ccccccccc}

    \toprule

    \textbf{Dataset} & $q$ & ARCNN~\cite{yu2016deep} & DnCNN~\cite{zhang2017beyond} & RNAN~\cite{zhang2019rnan} & FBCNN~\cite{Jiang_2021_ICCV} & DRUNet~\cite{zhang2021plug} & SwinIR~\cite{liang2021swinir} & ART~\cite{zhang2023accurate} & MambaIR~\cite{guo2024mambair} & \textbf{ATD-U} (ours) \\

    \midrule
    
    \multirow{4}{*}{\textbf{Classic5}~\cite{foi2007pointwise}}
    & 10 & 29.03/0.7929 & 29.40/0.8026 & 29.96/0.8178 & 30.12/0.8223 & 30.16/0.8234 & 30.27/0.8249 & \sotab{30.27}/\sotab{0.8258} & \sotab{30.27}/0.8256 & \sotaa{30.34}/\sotaa{0.8264} \\
    & 20 & 31.15/0.8517 & 31.63/0.8610 & 32.11/0.8693 & 32.31/0.8724 & 32.39/0.8734 & \sotab{32.52}/\sotab{0.8748} & -/-          & -/-          & \sotaa{32.58}/\sotaa{0.8755} \\
    & 30 & 32.51/0.8806 & 32.91/0.8861 & 33.38/0.8924 & 33.54/0.8943 & 33.59/0.8949 & 33.73/0.8961 & \sotab{33.74}/0.8964 & \sotab{33.74}/\sotab{0.8965} & \sotaa{33.80}/\sotaa{0.8967} \\
    & 40 & 33.32/0.8953 & 33.77/0.9003 & 34.27/0.9061 & 34.35/0.9070 & 34.41/0.9075 & 34.52/0.9082 & \sotab{34.55}/\sotab{0.9086} & 34.53/0.9084 & \sotaa{34.59}/\sotaa{0.9088} \\

    \midrule
    
    \multirow{4}{*}{\textbf{LIVE1}~\cite{sheikh2006statistical}}
    & 10 & 28.96/0.8076 & 29.19/0.8123 & 29.63/0.8239 & 29.75/0.8268 & 29.79/0.8278 & 29.86/0.8287 & \sotab{29.89}/0.8300 & 29.88/\sotab{0.8301} & \sotaa{29.91}/\sotaa{0.8304} \\ 
    & 20 & 31.29/0.8733 & 31.59/0.8802 & 32.03/0.8877 & 32.13/0.8893 & 32.17/0.8899 & \sotab{32.25}/\sotab{0.8909} & -/-          & -/-          & \sotaa{32.29}/\sotaa{0.8917} \\ 
    & 30 & 32.67/0.9043 & 32.98/0.9090 & 33.45/0.9149 & 33.54/0.9161 & 33.59/0.9166 & 33.69/0.9174 & \sotab{33.71}/0.9178 & \sotaa{33.72}/\sotab{0.9179} & \sotaa{33.72}/\sotaa{0.9180} \\ 
    & 40 & 33.63/0.9198 & 33.96/0.9247 & 34.47/0.9299 & 34.53/0.9307 & 34.58/0.9312 & 34.67/0.9317 & \sotab{34.69}/0.9317 & \sotaa{34.70}/\sotab{0.9320} & \sotab{34.69}/\sotaa{0.9321} \\
    
    \midrule
    
    \multirow{4}{*}{\textbf{Urban100}~\cite{Huang_2015_Urban100}}
    & 10 & -/-          & 28.54/0.8487 & 29.76/0.8723 & 30.15/0.8795 & 30.05/0.8772 & 30.55/0.8842 & \sotab{30.87}/\sotaa{0.8894} & \sotab{30.87}/0.8881 & \sotaa{30.88}/\sotab{0.8882} \\ 
    & 20 & -/-          & 31.01/0.9022 & 32.33/0.9179 & 32.66/0.9219 & 32.66/0.9216 & \sotab{33.12}/\sotab{0.9254} & -/-          & -/-           & \sotaa{33.32}/\sotaa{0.9265}\\ 
    & 30 & -/-          & 32.47/0.9248 & 33.83/0.9365 & 34.09/0.9392 & 34.13/0.9392 & 34.58/0.9417 & \sotaa{34.81}/\sotaa{0.9442} & \sotab{34.79}/\sotab{0.9427} & 34.76/0.9425 \\ 
    & 40 & -/-          & 33.49/0.9376 & 34.95/0.9476 & 35.08/0.9490 & 35.11/0.9491 & 35.50/0.9509 & \sotaa{35.73}/\sotaa{0.9553} & \sotab{35.70}/\sotab{0.9515} & 35.64/0.9514 \\

    \bottomrule
  \end{tabular}
  \end{center}

% \vspace{-0.7cm}
  
\end{table*}

\section{Conclusion}
In this work, we propose a novel image restoration model, termed Adaptive Token Dictionary.
Our model introduces a learnable token dictionary that captures representative image structures from the training data.
Building upon this dictionary, we design two key components: token dictionary cross-attention and adaptive category-based self-attention, complemented by several architectural modifications to exploit the embedded structural priors in the token dictionary.
Unlike existing window-based methods that rely on local self-similarity, our category-based partitioning strategy gathers similar features to form attention groups.
Each attention group comprises features features distributed across the entire image instead of local neighborhood, enabling effective global dependency modeling without compromising computational efficiency.
Furthermore, we leverage the category information extracted in TDCA to condition the feed-forward network, building the category-aware FFN to adaptively improve image features.
Extensive experimental results demonstrate that ATD and its multi-scale variant, ATD-U, outperform state-of-the-art methods across multiple image restoration tasks.
% \paragraph{Limitation.}
Moreover, we include a discussion on limitations and future work in the appendix.
These unexplored directions present promising avenues for future research.
We believe that this work paves the way for more effective and efficient solutions to global self-attention in image restoration tasks.

 {
    \small
    \bibliographystyle{IEEEtran}
    \sloppy
    \bibliography{IEEEfull}
}

\end{document}